\pdfoutput=1

\documentclass[11pt]{article}

\usepackage[final]{acl}

\usepackage{times}
\usepackage{latexsym}

\usepackage[T1]{fontenc} 

\usepackage[utf8]{inputenc} 
\usepackage[T1]{fontenc}    
\usepackage{hyperref}       
\usepackage{url}            
\usepackage{booktabs}       
\usepackage{amsfonts}       
\usepackage{nicefrac}       
\usepackage{microtype}      
\usepackage{xcolor}         
\usepackage{comment}        
\usepackage{graphicx}
\usepackage{caption}
\usepackage{appendix}
\usepackage{geometry}
\usepackage[most]{tcolorbox}
\usepackage{enumitem}
\usepackage{threeparttable}
\usepackage{mdframed} 
\usepackage{wrapfig}
\usepackage{subcaption}
\usepackage{makecell}
\usepackage{threeparttable}
\usepackage{amssymb}  

\usepackage{inconsolata}

\usepackage{graphicx}

\title{AfriMed-QA: A Pan-African, Multi-Specialty, Medical Question-Answering Benchmark Dataset}

\author{
  Tobi Olatunji$^{1,2,4}$  \thanks{These authors contributed equally to this work.}, Charles Nimo$^{2}$ \footnotemark[1], Abraham Owodunni$^{1,3}$ \footnotemark[1], \\
  \bf
  Tassallah Abdullahi$^{4,12}$, Emmanuel Ayodele$^1$, Mardhiyah Sanni$^{1,4}$, 
  Chinemelu Aka$^1$, \\
  \bf
  Folafunmi Omofoye$^{4}$, Foutse Yuehgoh$^4$, Timothy Faniran$^4$, Bonaventure F. P. Dossou$^{4,10, 13}$, \\
  \bf
  Moshood Yekini$^4$,
  Jonas Kemp$^6$, Katherine Heller$^6$, Jude Chidubem Omeke$^4$, \\
  \bf
  Chidi Asuzu MD$^4$,
  Naome A. Etori$^{4,11}$, Aimérou Ndiaye$^5$, Ifeoma Okoh$^5$, Evans Doe Ocansey$^5$, \\
  \bf
  Wendy Kinara$^{7}$, Michael Best$^2$, Irfan Essa$^{2,6}$, Stephen Edward Moore$^{8}$, \\
  \bf
  Chris Fourie$^{9}$, Mercy Nyamewaa Asiedu$^{6}$ \thanks{ Senior authors.}\\
  $^1$Intron, $^2$Georgia Institute of Technology, $^3$The Ohio State University, $^4$BioRAMP, \\  
  $^5$Masakhane, $^6$Google Research, $^7$Kenyatta University, $^8$University of Cape Coast, \\ 
  $^9$SisonkeBiotik, $^{10}$MILA Quebec, $^{11}$University of Minnesota, $^{12}$Brown University $^{13}$McGill University \\
  \texttt{tobi@intron.io}
}

\begin{document}
\maketitle

\begin{abstract}
Recent advancements in large language model (LLM) performance on medical multiple-choice question (MCQ) benchmarks have stimulated interest from healthcare providers and patients globally. Particularly in low-and-middle-income countries (LMICs) facing acute physician shortages and lack of specialists, LLMs offer a potentially scalable pathway to enhance healthcare access and reduce costs. However, their effectiveness in the Global South, especially across the African continent, remains to be established. In this work, we introduce AfriMed-QA, the first large-scale Pan-African English multi-specialty medical Question-Answering (QA) dataset, 15,000 questions (open and closed-ended) 
sourced from over 60 medical schools across 16 countries, covering 32 medical specialties. We further evaluate 30 LLMs across multiple axes including correctness and demographic bias. Our findings show significant performance variation across specialties and geographies, MCQ performance clearly lags USMLE (MedQA). We find that  biomedical LLMs underperform general models and smaller edge-friendly LLMs struggle to achieve a passing score. Interestingly, human evaluations show a consistent consumer preference for LLM answers and explanations when compared with clinician answers.

\end{abstract}

\section{Introduction}

\begin{figure*}[t]
    \centering
    \includegraphics[width=0.9\linewidth]{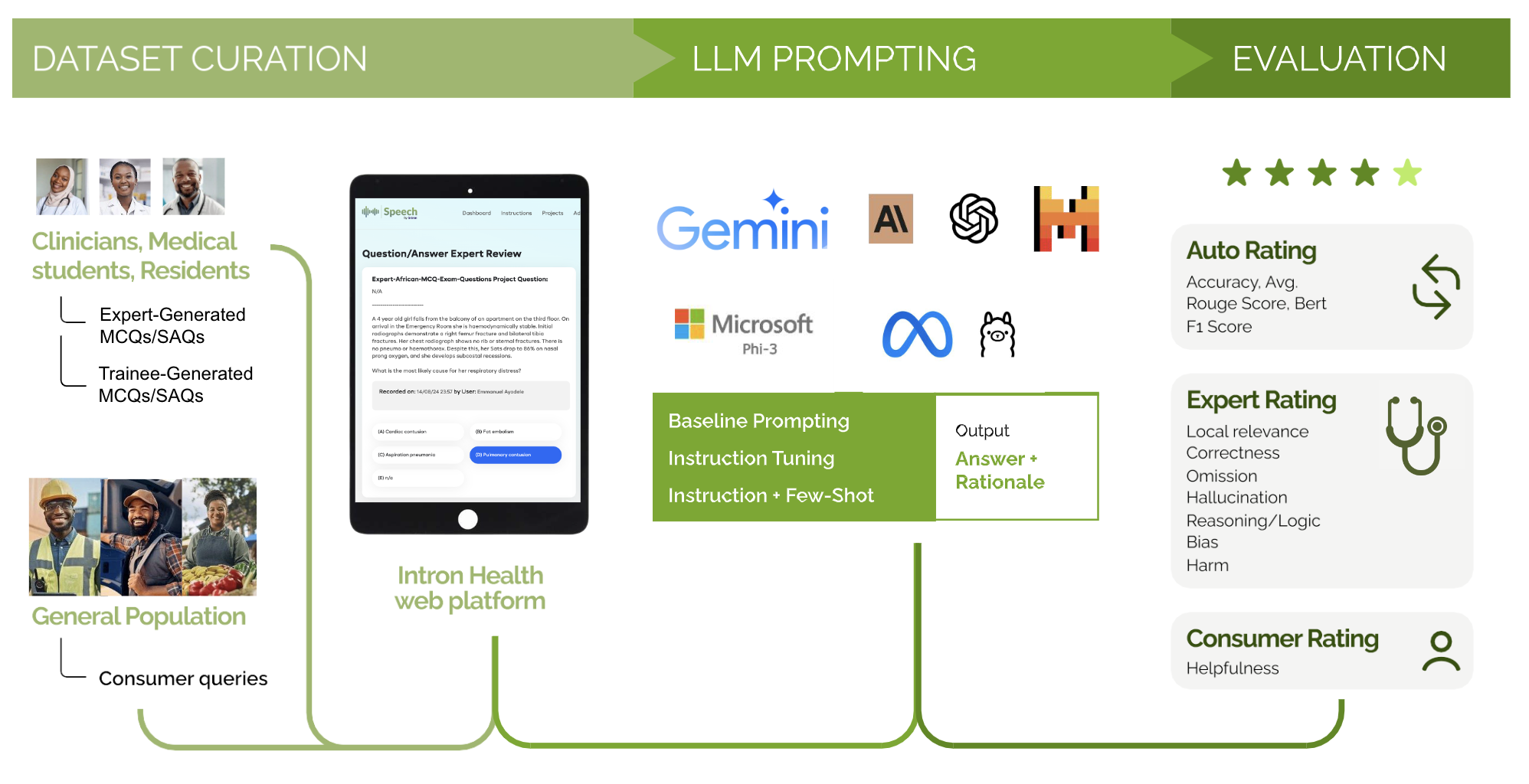}
    \caption{Methodology Overview} 
        \label{fig:overview}
    \end{figure*}

\begin{table*}[t]
\caption{Comparative analysis of the AfriMed-QA dataset with health datasets.}
\centering
\resizebox{\textwidth}{!}{%
\begin{tabular}{lcccccccc}
\toprule
\textbf{Feature} & 
\textbf{AfriMed-QA} & 
\makecell{\textbf{BioASQ}\\ \cite{422}} & 
\makecell{\textbf{MedQA-SWE}\\ \cite{hertzberg-lokrantz-2024-medqa-swe}} & 
\makecell{\textbf{MedQA}\\ \cite{Jin2020WhatDD}} & 
\makecell{\textbf{MedMCQA}\\ \cite{pmlr-v174-pal22a}} & 
\makecell{\textbf{PubMedQA}\\ \cite{jin-etal-2019-pubmedqa}} & 
\makecell{\textbf{MMLU}\\ \cite{Hendrycks2020MeasuringMM}} & 
\makecell{\textbf{HealthSearchQA}\\ \cite{Singhal2022LargeLM}} \\
\midrule
\textbf{Dataset Size} & 15,275 & 4,721 & 3,180 & 12,723 & 193,155 & 1,000 & 15,908 & 3,173 \\
\textbf{Question Types} & MCQ, SAQ, & yes/no, Factoid, & MCQ & MCQ & MCQ & Yes/No/maybe, & MCQ & Consumer Queries \\
\textbf & Consumer Queries & List, Summary &  &  &  & Factoid, List &  &  \\
\textbf{Clinical Scenarios} & $\checkmark$ & $\times$ & $\times$ & $\checkmark$ & $\times$ & $\times$ & $\times$ & $\checkmark$ \\
\textbf{Answer Options} & 2-5 (MCQs) & $\times$ & 5 (MCQs) & 4 (MCQs) & 4 (MCQs) & $\times$ & 4 (MCQs) & $\times$ \\
\textbf{Correct Answers} & $\checkmark$ & $\checkmark$ & $\checkmark$ & $\checkmark$ & $\checkmark$ & $\checkmark$ & $\checkmark$ & $\checkmark$ \\
\textbf{Answer Rationale} & $\checkmark$ & $\checkmark$ & $\times$ & $\checkmark$ & $\checkmark$ & $\checkmark$ & $\times$ & $\checkmark$ \\
\textbf{Question Source} & $\checkmark$ & $\checkmark$ & $\times$ & $\times$ & $\times$ & $\checkmark$ & $\times$ & $\times$ \\
\bottomrule
\end{tabular}}
\label{tab:comparison}
\end{table*}

Large language models (LLMs) have gained popularity in specialized domains such as finance \cite{wu2023bloomberggpt}, medicine \cite{singhal2022large}, and climate \cite{thulke2024climategpt}. In the medical field, LLMs like Med-PaLM \cite{singhal2022large}, PMC-LLaMA \cite{wu2023pmc}, and GPT-4 \cite{achiam2023gpt} have shown impressive performance in tasks such as summarizing clinical notes and answering medical questions with high accuracy \cite{liu2024exploring, eriksen2023use, singh2024learning, liu2024benchmarking}. Especially in low-resource settings, these models have the potential to improve clinician productivity, accessibility, operational efficiency, and enable multilingual clinical decision support \cite{yang2023large, gangavarapu2023llms}. \\
Despite their success on existing medical benchmarks, it is uncertain whether these models generalize to tasks involving linguistic variations, even within English, localized cultural contexts, and region-specific medical knowledge, highlighting the need for more diverse benchmark datasets. Current evaluations rely on publicly available digital resources \cite{kung2023performance, jin-etal-2019-pubmedqa, jin2021disease}, but these may not translate to out-of-distribution datasets, such as those from African countries.\\
To address this gap, we introduce AfriMed-QA,
\footnotetext[1]{Available at: \\ \href{https://huggingface.co/datasets/intronhealth/afrimedqa_v2}{https://huggingface.co/datasets/intronhealth/afrimedqa\_v2}}
a dataset of 15,275 English, clinically diverse questions and answers, 4,000+ expert multiple-choice questions (MCQs) with answers, over 1,200 open-ended short answer (SAQs) with long-form answers, and 10,000 consumer queries (CQ), to rigorously assess LLM performance for correctness and demographic bias. We evaluate 30 large, small, open, closed, biomedical, and general LLMs using quantitative and qualitative approaches. While the development of the dataset is still in progress, this work establishes a foundation for acquiring diverse and representative health benchmark datasets across LMICs. The dataset is released under a CC-BY-NC-SA 4.0 license.  


\section{Related Work}
\subsection{Medical Domain Benchmark Datasets for LLMs}
The \textit{Open Medical LLM Leaderboard} \cite{Medical-LLM} tracks, ranks, and evaluates LLM performance on medical question-answering tasks across diverse datasets, including MedQA (USMLE) \cite{jin2021disease}, PubMedQA \cite{jin-etal-2019-pubmedqa}, MedMCQA \cite{Pal2022MedMCQAA}, and subsets of MMLU \cite{Hendrycks2020MeasuringMM}. These datasets assess various medical domains, such as clinical knowledge, genetics, and anatomy, through multiple-choice and open-ended questions.
 CMExam \cite{NEURIPS2023_a48ad12d} from the Chinese National Medical Licensing Examination offers MCQs with detailed annotations, while Q-Pain \cite{Loge2021QPainAQ}, Medication QA \cite{Abacha2019BridgingTG}, LiveQA \cite{qianying-etal-2020-liveqa}, MultiMedQA \cite{Singhal2022LargeLM}, and EquityMedQA \cite{pfohl2024toolbox} cover various medical QA challenges. Table \ref{tab:comparison} compares AfriMed-QA to other medical QA benchmarks.

\subsection{Evaluating LLMs for Health-Specific Tasks}

There have been several studies evaluating LLMs for medical standardized exams and for various clinical tasks using approaches like zero-shot, finetuning, developing benchmarking metrics and running human evalutations  \cite{JAHAN2024108189, singhal2022large,Fleming_2024,umapathi2023med, chen2024benchmarking}. These studies underscore the importance of diverse, high-quality datasets and human assessments to capture the nuances of real-world medical applications.

\citet{REDDY2023101304} proposed the TEHAI framework to assess the translational and governance aspects of LLMs in healthcare, emphasizing contextual relevance, safety, ethical considerations, and efficiency. 

Our evaluation extends TEHAI and work from \citet{singhal2022large}, and \citet{pfohl2024toolbox} by incorporating specialty and region-specific dimensions, ensuring a comprehensive assessment of LLMs. Metrics such as local expertise, harmfulness, and bias are included to cover ethical dimensions and governance. By addressing correctness, omission, hallucination, and reasonableness, we thoroughly evaluate AI systems in healthcare.

The AfriMed-QA dataset aims to (1) integrate geo-culturally diverse datasets, specifically those from African LMICs that have historically relied on paper-based records, and local health data and are underrepresented in LLM training and evaluation; and (2) expand healthcare LLM benchmark datasets to include African consumer/patient-based queries. This enables LLM training and evaluation on a broad spectrum of medical data, creating more robust, inclusive, and practical AI solutions for Africa-centric applications.

\section{AfriMed-QA Dataset}

We introduce AfriMed-QA, the first large-scale Pan-African multispecialty medical Question-Answer dataset designed to evaluate and develop equitable and effective LLMs for African healthcare. Figure \ref{fig:overview} details our methodology and data collection process. 

\begin{table}[!htpb]
\small
\centering
\begin{tabular}{@{}lccc@{}}
\toprule
Question Tier  &  Type    & Count   & Total  \\
\midrule
Expert  & MCQ    & 3,910   & 4,269  \\
  & SAQ    & 359   &   \\
\midrule
Crowdsourced  & MCQ    & 129   & 11,006  \\
  & SAQ    & 877   &   \\
  & CQ    & 10,000   &    \\
\midrule
Total   &     &   &   \textbf{15,275} \\
\bottomrule
\end{tabular}
\caption{Dataset statistics}
\label{tab:data-stats}
\end{table}

\begin{figure}[t!]
    \centering   
    \includegraphics[width=0.9\linewidth]{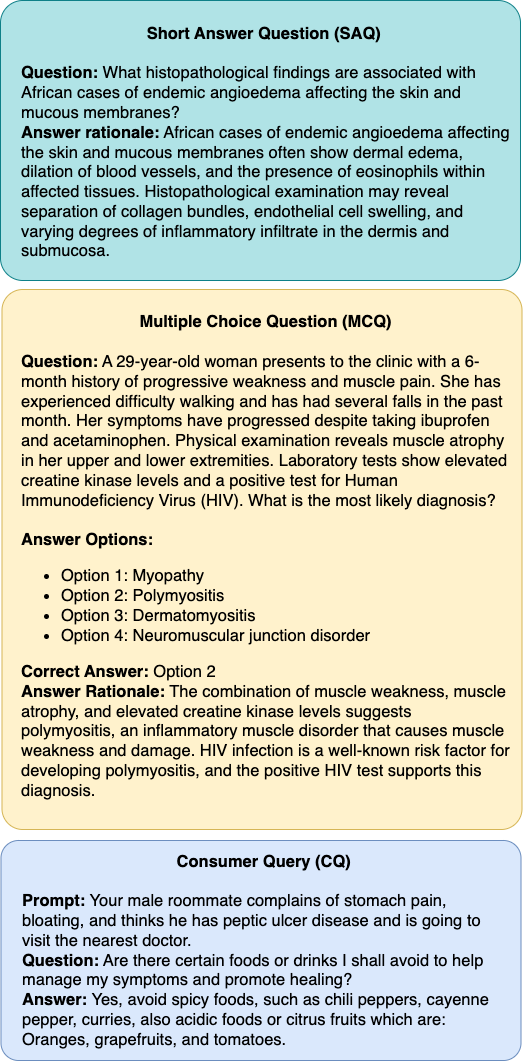}
    \caption{Question Samples for multiple choice questions (MCQs), Short Answer Questions (SAQs), and Consumer Queries} 
        \label{fig:sample_ques}
    \end{figure}

Table \ref{tab:data-stats} and \ref{tab:country-counts} shows dataset statistics. Multiple-choice (MCQ) professional medical exam questions include a question, 2-5 alternative answer options, the correct answer(s), and the rationale for the correct answer (answer option count distribution is shown in Appendix Tab \ref{tab:ans-opt-distr}). Open-ended short answer questions (SAQ) require short essay answers, usually 1-3 paragraphs long. Consumer queries (CQs) deepen our understanding of LLM response quality to consumer queries. To maximize the diversity of consumer questions, we leveraged a curated list of 472 medical conditions, symptoms, and patient complaints common in African communities across 32 specialties, to create culturally appropriate prompts that help elicit diverse questions from clinical and non-clinical crowd workers. Example MCQ, SAQ, and CQ questions, as well as CQ human prompt templates are shown in Figure \ref{fig:sample_ques}.

This dataset was crowd-sourced from 621 contributors (Female 55.56\%, Male 44.44\%) from over 60 medical schools across 16 countries (Table \ref{tab:country-counts}), covering 32 medical specialties including Obstetrics \& Gynecology, Neurosurgery, Internal Medicine, Emergency Medicine, Medical genetics, Infectious Disease, and others. Appendix Table \ref{table:combined_specialty_count} shows the Specialty distribution. Human answers and explanations are provided for 5,444 questions.  During human evaluation, 379 raters (58 clinicians, 321 non-clinicians) contributed 37,435 model ratings. 


\subsection{Data Collection}
We adapted a web-based platform previously developed by Intron Health 
\footnotetext[2]{Intron Health's biomedical crowd-sourcing platform \url{https://speech.intron.health}}
to crowd-source accented and multilingual clinical speech data at scale across Africa \cite{olatunji2023afrispeech}. 
Custom user interfaces were developed to collect each question type, for quality reviews, and for blind human evaluation of LLM responses. The MCQ User Interface and other details about the data collection tool can be found in Appendix Figure \ref{fig:intron-ui-tool}. 

\subsection{Contributor Recruitment and Instructions for Data Collection and Evaluation} \label{sec:contributors}
\textbf{Contributors:} Medical trainees and clinicians were recruited to contribute questions through referrals from existing Intron Health's web-based platform contributors, medical associations, and via social media. Experts (Professors) were recruited from Medical Schools in 5 countries as shown in Appendix Table \ref{fig:country}. Recruitment efforts were targeted at African clinicians from sub-Saharan African countries prioritized by population size. 
To maximize geographic representation, each contributor was limited to a maximum of 300 questions and answers. Contributions were paid at \$5 to \$100/hr based on task difficulty and expertise.

 \textbf{Instructions:} For MCQs and SAQs, clinician contributors were instructed to 1) input question, answer options, correct answer(s), rationale/explanation for correct answer, and question metadata into the interface. Experts provided MCQ answers with no explanations. 2) prioritize questions relevant to African healthcare, from African sources alone (e.g. no USMLE prep questions). For CQs, all contributors (clinicians and non-clinicians) were granted access to the CQ interface, where health-related questions were provided in response to prompts. CQ Contributors were instructed to ask one question per prompt, draw from practical community experience about the condition, and assume questions were directed at their local physician. Clinician contributors were then granted privileged access to a dedicated interface to provide human answers to consumer queries.

\textbf{Human Evaluations:} Consumers provided ratings for LLM responses to CQs on relevance, helpfulness, and local context, but NOT correctness. Due to the higher expertise required, only confirmed clinicians were granted access to projects rating MCQ, SAQ, and CQ answers. Clinician status was confirmed through submitted credentials and background checks reviewing publicly available information about them. Raters were randomly assigned (double-blind) to the answer source (human or LLM). More details on human evaluations are provided in Section \ref{sec:human-eval}

\textbf{Quality Review:} We utilized the contributor quality review process described in \cite{olatunji2023afrispeech}. Contributors were rigorously evaluated by a team of clinicians. Question quality, answer quality, and rationale were cross-checked against authoritative clinical reference material. Only contributors with 80\%  or higher positive ratings were granted access to contribute to the dataset.

\section{Approach}

\subsection{LLM Selection}
We evaluate 30 LLMs (Table \ref{models_perf}) including open-source and proprietary LLMs, general-purpose and biomedical LLMs, mixture-of-experts, and models of varying sizes from 3B to over 540B parameters. Of these, 13 are proprietary while 17 are open-source. We evaluate LLMs like Phi-3 \cite{abdin2024phi}, GPT-4 \cite{achiam2023gpt}, MedLM, Claude 3 \cite{anthropic_claude_2023}, OpenBioLLM \cite{OpenBioLLMs}, Gemini \cite{team2023gemini}, Meditron \cite{chen2023meditron}, and MedLlama \cite{wu2023pmc}.

\subsection{Quantitative Evaluation}

For multiple-choice tasks, accuracy is measured by comparing LLM's single-letter answer choice [A,B,C,D,E] with the reference. For open-ended questions, we evaluate semantic similarity using BERTScore \cite{bert-score} and QuestEval \cite{scialom2021questevalsummarizationasksfactbased}, both comparing the generated response from the language model against a reference answer, and sentence-level structural overlap using ROUGE-Lsum \cite{lin-2004-rouge}, which likewise compares the generated response against its reference.

\begin{figure*}[t]
    \centering
    \includegraphics[width=1\textwidth]{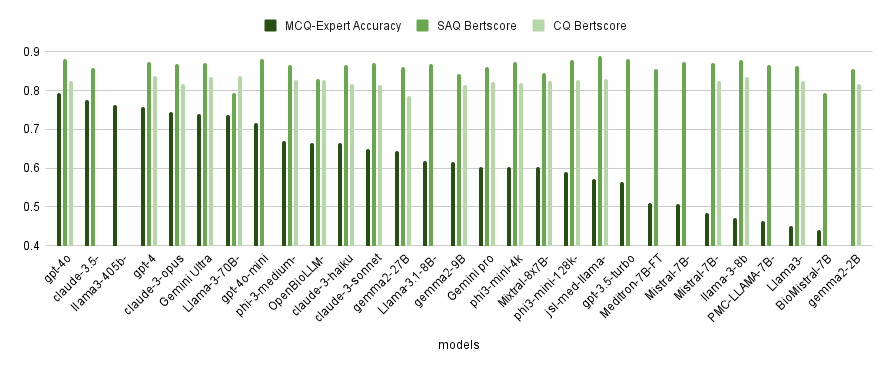}
    \caption{AfriMedQA: Expert-MCQ accuracy, SAQ, and CQ Bertscore, sorted by Expert-MCQ Accuracy}
    \label{fig:main-results}
\end{figure*}%

\begin{figure*}[h!]
    \centering
    \begin{subfigure}[t]{0.4\textwidth}
        \centering
        \includegraphics[width=\textwidth]{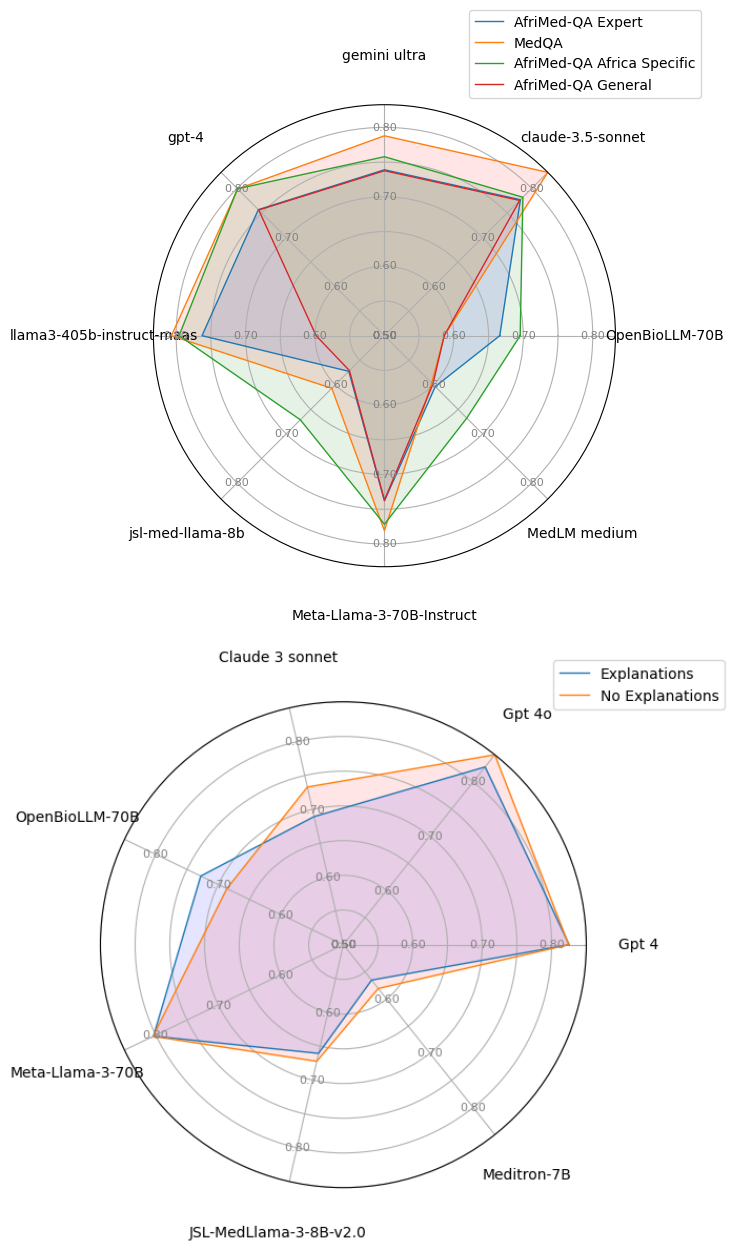}
        \caption{Top: MedQA vs AfriMedQA MCQ Accuracy. Bottom: Effect of Explanations}
        \label{fig:medqa}
    \end{subfigure}%
    \hfill
    \begin{subfigure}[t]{0.6\textwidth}
        \centering
        \includegraphics[width=\textwidth]{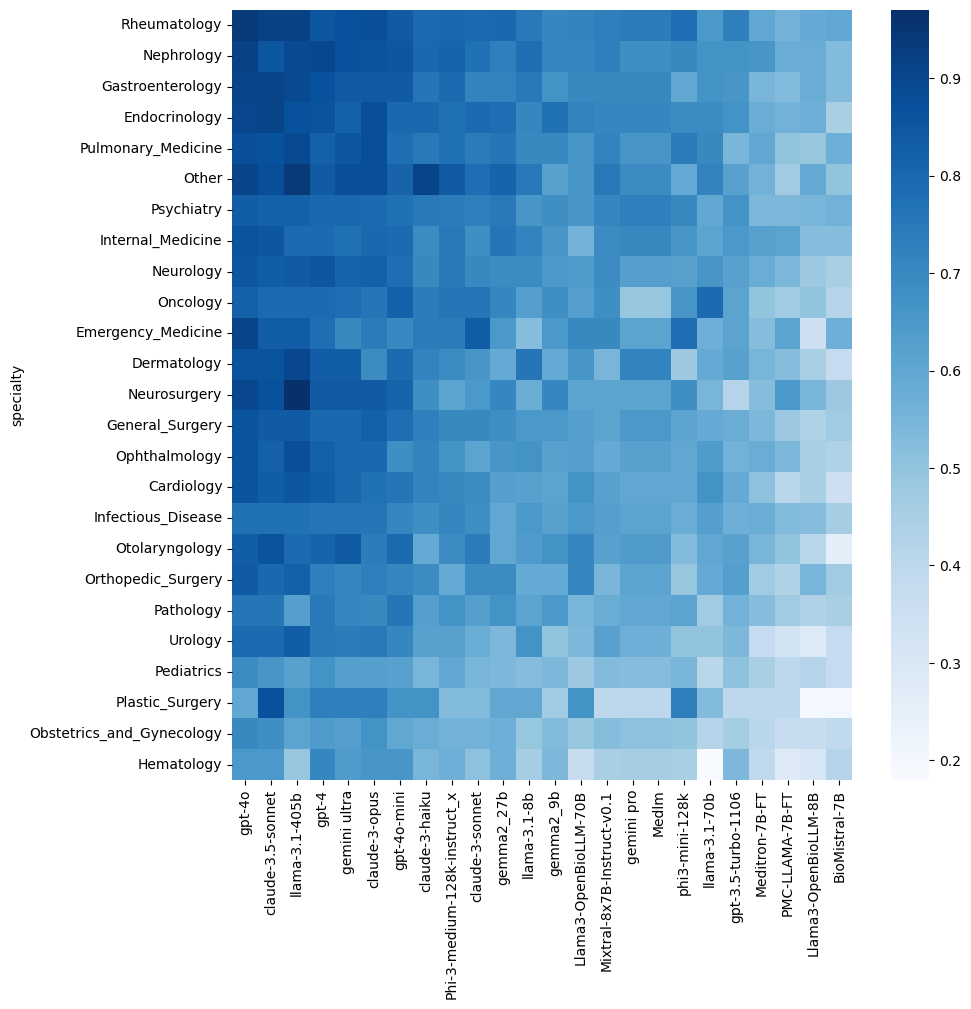}
        \caption{MCQ accuracy by specialty}
        \label{fig:specialty-accuracy}
    \end{subfigure}
    
    \caption{Breakdown of MCQ accuracy by specialty and dataset}
    \label{fig:mcq_perf}
\end{figure*}

\begin{figure}[t!]
    \centering
    \rotatebox[origin=c]{270}{\includegraphics[width=0.8\linewidth]{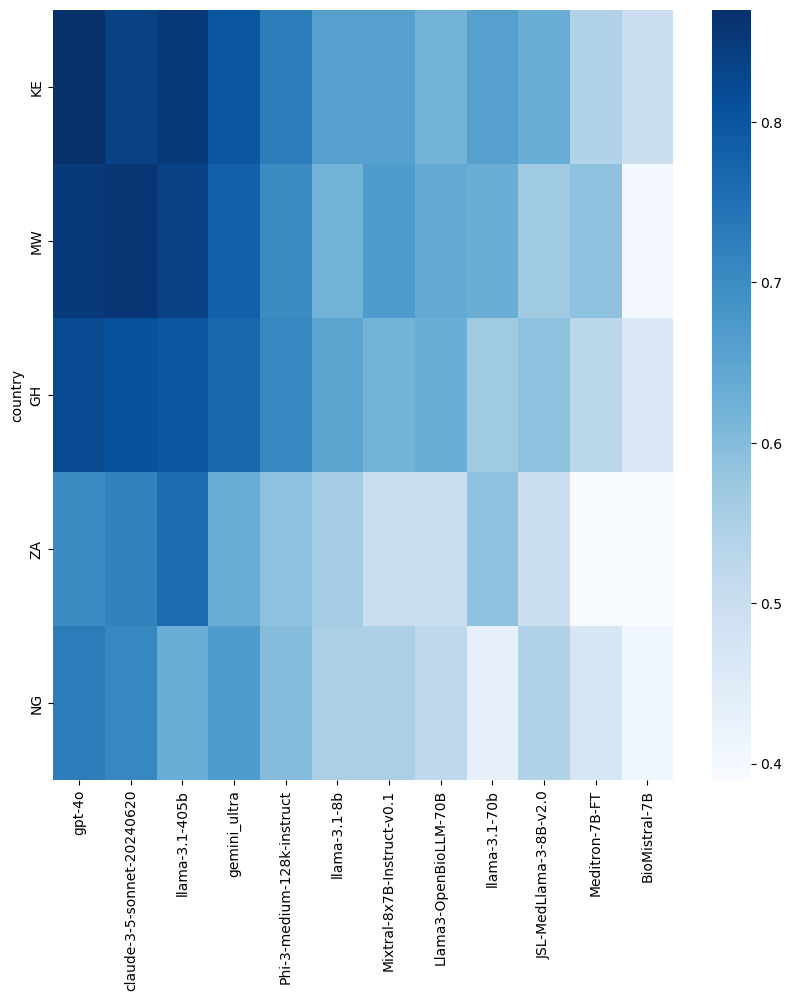}}
    \caption{MCQ accuracy by country}
    \label{fig:mcq-accuracy-by-country}
\end{figure}

\subsection{Qualitative: Human Evaluations and Evaluation Axes} 
\label{sec:human-eval}
LLM responses to a fixed subset of questions (n=3000, randomly sampled) were sent out for human evaluation on the Intron Health crowd-sourcing platform. 
Adapting the evaluation axes in \cite{singhal2023towards}, we collected human evaluations in two categories: (1) \textbf{Non-clinicians} were instructed to provide ratings to CQ LLM responses to determine if answers were relevant, helpful, and localized; (2) \textbf{Clinicians} were instructed to provide ratings to the LLM's MCQ, SAQ, and CQ responses to determine if answers were correct and localized, if omissions or hallucinations were present, and if potential for harm exists. Evaluation axes and exact instructions are detailed in Appendix section \ref{sec:human-eval-axes}.

Ratings were on a 5-point scale representing the extent to which the criteria were met. 1 represents "No" or "completely absent", and 5 represents "Yes" or "Absolutely present". Raters were blinded to the answer source (model name or human). Each rater evaluated multiple LLMs or human answers in random blind sequence.


\subsection{Experiment Setup} 
\label{sec:setup} 
 Checkpoints for open-source models were sourced from HuggingFace and Google's Vertex AI Studio, while proprietary models were accessed through developers' API using default hyperparameters. More details about the hyper-parameters used in this study are available in the Appendix in Table \ref{tab:hyperparameters}. Aligning with our community participatory design approach, experiments by multiple collaborators ran on various hardware types including 1 of either NVIDIA L4, NVIDIA T4, or A100.

\section{Results}

\subsection{Benchmark evaluations}
AfriMed-QA MCQ, SAQ, and CQ evaluation results are shown in Figure \ref{fig:main-results}. Accuracy ranges from 0.17 (Gemma-2B, the smallest LLM in this study) to 0.79 (GPT-4o). GPT-4o, Claude-3.5-sonnet and Llama3-405b are the top 3 most accurate LLMs (granular details are provided in Appendix Table \ref{models_perf}). Using Base Prompts (Appendix \ref{fig:question_instructions}), we assess LLM performance by country, (Fig \ref{fig:mcq-accuracy-by-country}), specialty (Fig \ref{fig:specialty-accuracy}), data subset (Fig \ref{fig:medqa}), and with or without explanations (Fig \ref{fig:medqa}). 

\subsection{Human evaluations}
\begin{figure*}[htbp]
    \centering

    \begin{minipage}{\textwidth}
        \centering
        \begin{subfigure}[b]{0.48\textwidth}
            \centering
            \includegraphics[width=\textwidth]{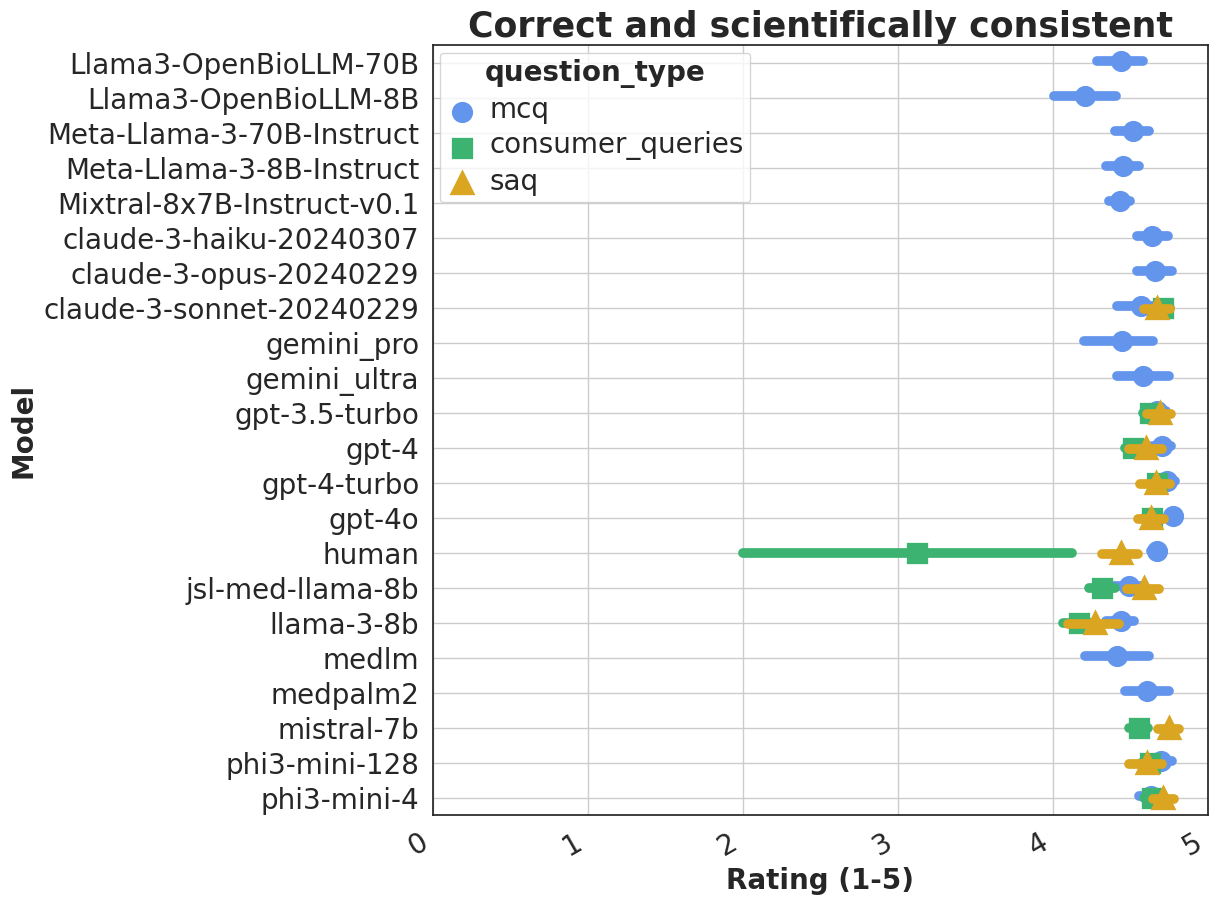}
            \caption{Correctness}
            \label{fig:corr_clin}
        \end{subfigure}
        \hfill
        \begin{subfigure}[b]{0.48\textwidth}
            \centering
            \includegraphics[width=\textwidth]{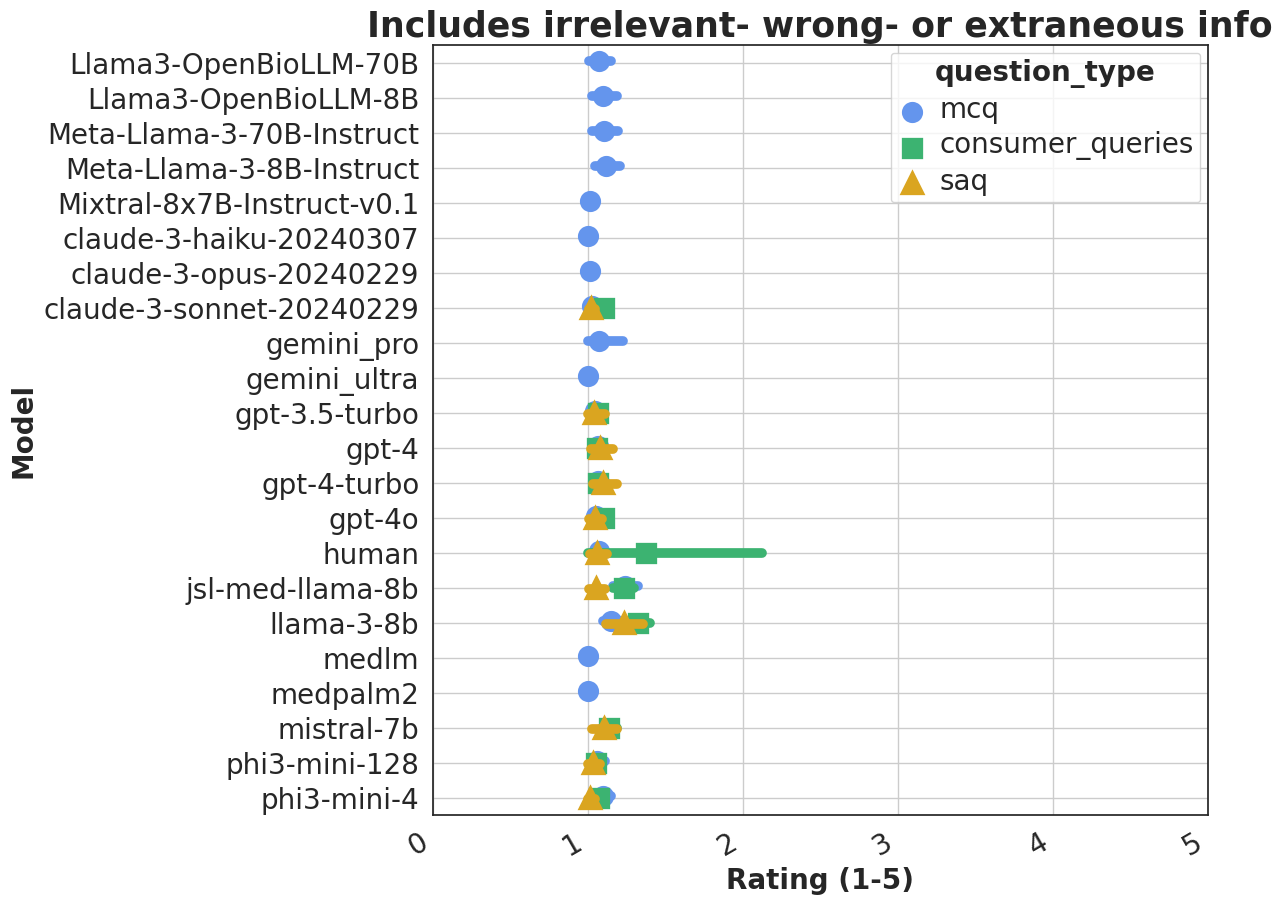}
            \caption{Irrelevant information}
            \label{fig:irre_clin}
        \end{subfigure}
    \end{minipage}

    \vspace{0.3cm}
    \begin{minipage}{\textwidth}
        \centering
        \begin{subfigure}[b]{0.48\textwidth}
            \centering
            \includegraphics[width=\textwidth]{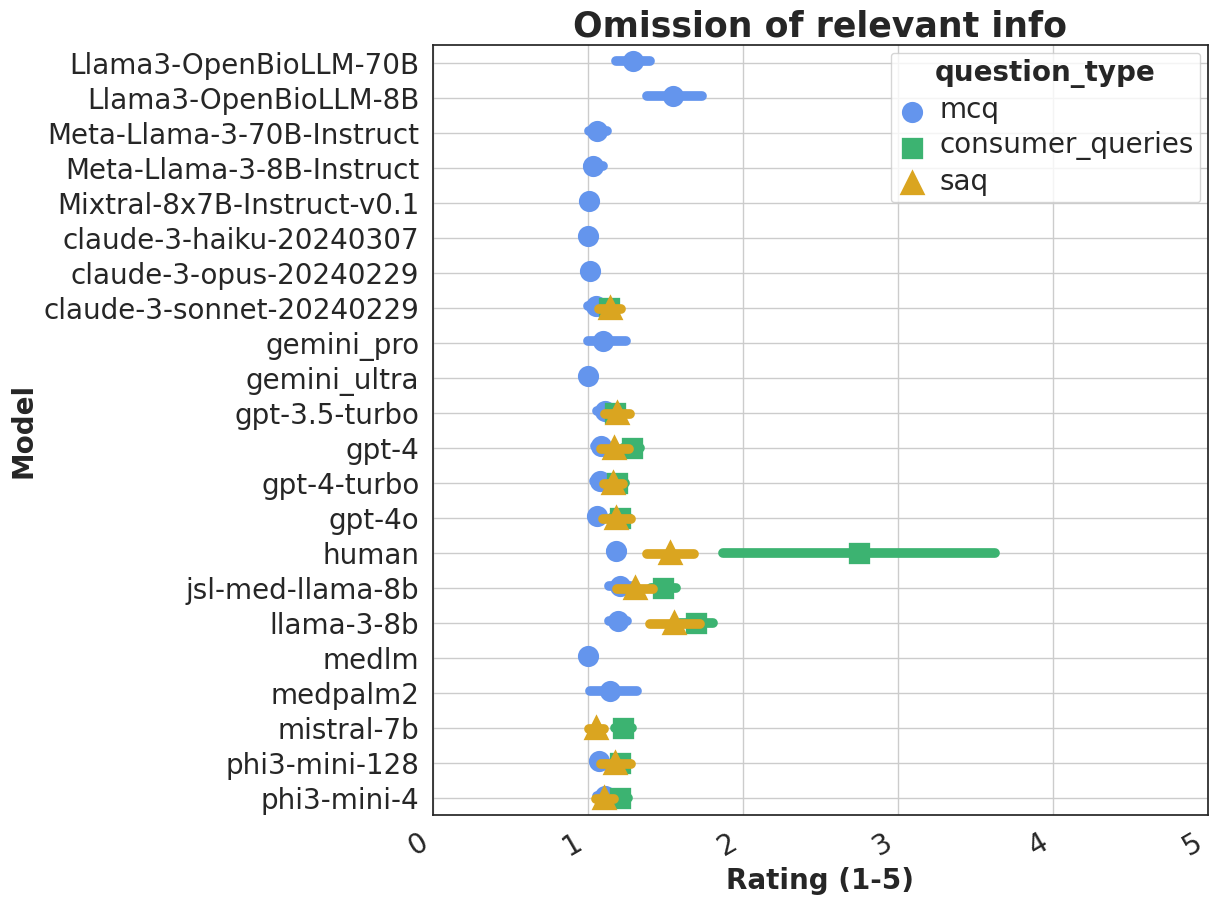}
            \caption{Omission of relevant information}
            \label{fig:omis_clin}
        \end{subfigure}
        \hfill
        \begin{subfigure}[b]{0.48\textwidth}
            \centering
            \includegraphics[width=\textwidth]{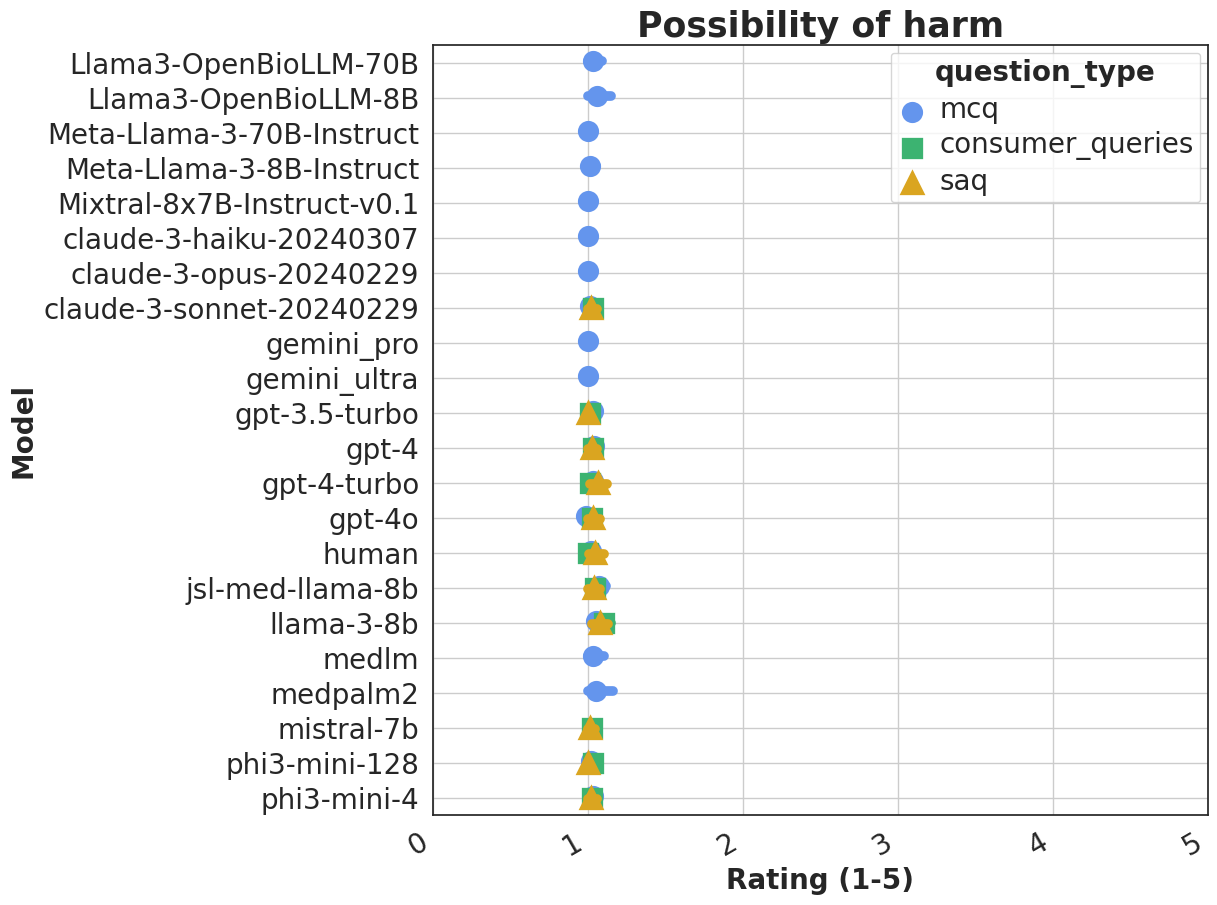}
            \caption{Possibility of harm}
            \label{fig:harm_clin}
        \end{subfigure}
    \end{minipage}

    \vspace{0.3cm}
    \begin{minipage}{\textwidth}
        \centering
        \begin{subfigure}[b]{\textwidth}
            \centering
            \includegraphics[width=\textwidth]{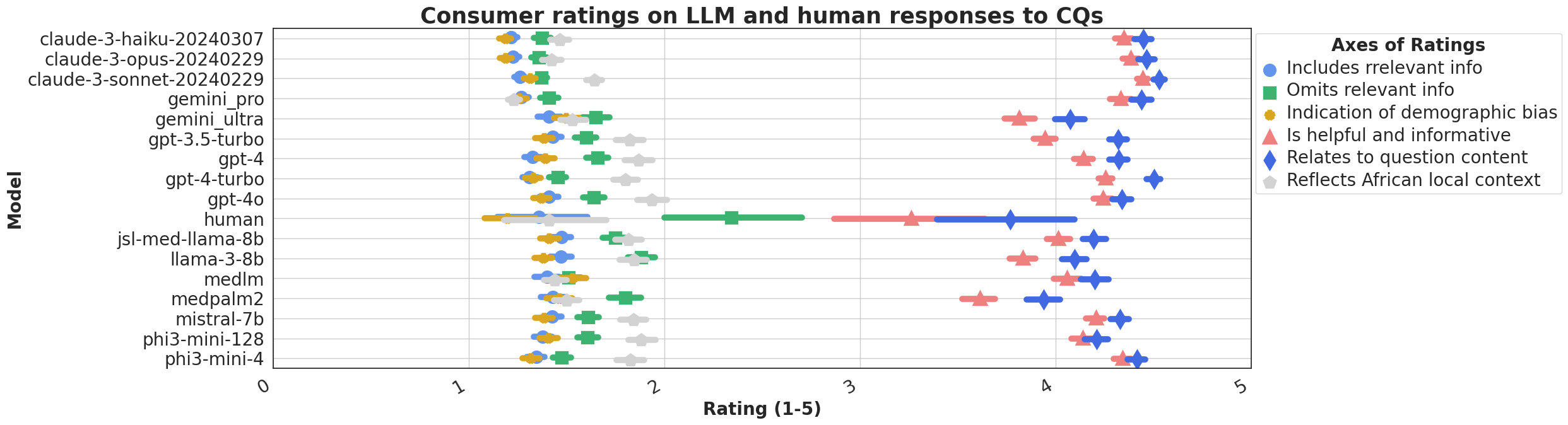}
            \caption{Consumer ratings of CQs}
            \label{fig:consumer}
        \end{subfigure}
    \end{minipage}

    \caption{Clinician and consumer blind evaluations of human and LLM answers showing mean ratings and confidence intervals across various axes. 
    }
    \label{fig:human_eval_clin}
\end{figure*}

Figure \ref{fig:human_eval_clin} shows results from clinician and non-clinician human evaluators on the dataset across various axes for LLM and Human responses. Figure \ref{fig:consumer} shows non-clinician ratings of responses to consumer queries from LLMs and humans (blinded). Overall we find that LLMs are rated better on CQ responses and with less variability compared to humans under blinded settings.  

\section{Discussion}



The experimental results of our study on AfriMed-QA reveal several key insights and trends.

\subsection{The dominance of large models may be unfavorable for low-resource settings}
Figure \ref{fig:main-results} shows a wide range of LLM accuracies on Expert MCQs with the best models (large, 100B-2T, proprietary) scoring over 75\%, smaller models (under 10B) clustering in the 40-60\% range, and medium models (11B-70B) scoring between 60 and 75\% (details in Appendix Table \ref{models_perf}). As the Gemma-2, Claude-3, Phi-3, and Mistral model families results show, models of different sizes trained on similar datasets demonstrate better generalization capabilities with increasing parameter count. Mixture-of-Expert (MoE) models Mistral-8x7B outperforms its biomedical and general 7B variants by a wide-margin further confirming the correlation between model capacity and performance. Blind clinician evaluation of LLM answers to open and closed-ended questions (Figure \ref{fig:human_eval_clin}) are consistent with this trend, showing larger models are more correct and less susceptible to hallucinations and omissions than small models. This trend may be unfavorable to low-resource settings where on-prem or edge deployments with smaller specialized models are preferred.

\subsection{Localization is still a challenge}
Figure \ref{fig:medqa} (top left) reveals an unmistakable performance gap between USMLE (MedQA) (orange line) and AfriMed-QA Expert MCQs (blue line), with proprietary GPT-4o, Claude-3.5-sonnet and Gemma-2B showing an 8.86, 5.57, and 15.5 point drop in performance (Appendix Table \ref{medqa_vs_expert_mcq}) which may indicate bias attributable to their training data distribution. Figure \ref{fig:medqa} further shows differences in question difficulty by source. Since MCQs that mention African cities or locations (e.g. A 50-year old patient with recent travel to Uganda...) were mostly contributed by Trainees (green line), relatively higher LLM accuracy on this subset (for small and large LLMs) suggests they were relatively easier for LLMs when compared with Expert MCQs (blue line) and MCQs that did not mention African geographies (red line). This finding is counter-intuitive and requires further investigation.

\subsection{Understanding the domain shift}
Although it seems intuitive that medical knowledge is universal, our findings highlight key reasons for regional differences in LLM Performance. Questions require not only general biomedical knowledge but also contextual understanding of African-specific disease patterns, local health challenges, and socio-cultural factors. For example, children in specific communities may be required to take vaccines (e.g. HPV) earlier or later based on logistical access challenges. Diseases may also involve region-specific appearance of symptoms (e.g. appropriate color for skin lesion) and  clinical presentation (e.g. cancer patients typically seek professional medical care much later than Western countries, affecting how they are managed on the first visit), and management may reflect available medications, treatment options, and diagnostic equipment. Although there is significant medical knowledge overlap globally, regional variations exist, necessitating physician board certification in each country. Such local variations make questions more challenging for LLMs trained mainly on Western medical data.

\subsection{Evidence of progress in LLM reasoning abilities}
Performance of the GPT model series (3.5, 4, and 4o in Fig \ref{fig:main-results}) show an interesting temporal trend with newer models scoring significantly higher on both MedQA and AfriMed-QA showing strong evidence of progress with LLM reasoning abilities not attributable to question memorization since the novel AfriMed-QA questions were not part of any GPT version's training data.

\subsection{Domain-specific biomedical LLMs still struggle}
Figure \ref{fig:main-results} shows general models outperform and generalize better than biomedical models of similar size (8B and 70B). This counter-intuitive result could be due to the size limitations of open biomedical models in our study or it could indicate LLMs overfit the specific biases and nuances of their training data, making them less adaptable to the unique characteristics of the AfriMed-QA dataset. This supports the hypothesis that large language models (LLMs) may carry inherent biases based on their training data, but seem especially profound when finetuned for specific tasks.

\subsection{Specialties must select LLMs with caution}
Figure \ref{fig:specialty-accuracy} shows a clear top-down, left-right trend indicating which LLMs are more reliable in certain specialties. Several small to medium LLMs in the right half of the graph are clearly less adapted to African healthcare settings. While larger closed LLMs seem to generalize across specialties, our results reveal that LLMs perform better on Medical Specialties (Rheumatology, Nephrology, Gastroenterology, Endocrinology, Pulmonary, etc) when compared with other specialties like Surgery, Pathology, Pediatrics, Infectious Diseases, and Obstetrics \& Gynecology that are very important in LMICs, a striking trend that requires further investigation.

\subsection{Performance variation by country}
Figure \ref{fig:mcq-accuracy-by-country} shows a clear difference in difficulty of expert question from South Africa and Nigeria for LLMs. This requires further investigation but could be a result of differences in specialty distributions of expert MCQs per country. For example, expert questions from South Africa are dominated by Pediatrics, a difficult specialty for LLMs as shown in Figure \ref{fig:specialty-accuracy} while a significant number of Pathology and Obstetrics and Gynecology questions come from Nigeria. Country-Specialty counts are detailed in Appendix Tables \ref{tab:country-specialties}, \ref{mcq_acc_countries}, and \ref{table:country_count}.

\subsection{Limitations of automated metrics for evaluating SAQ and CQ answers}



As shown by the narrow range of BERTScore values in Appendix Table~\ref{models_perf} (all LLMs cluster between 0.86 and 0.89), its utility in this context is limited—a problem also identified in the literature \cite{hanna2021fine, celikyilmaz2020evaluation, zhang2024rouge}. Appendix Table~\ref{models_perf} also reports ROUGE-Lsum and QuestEval scores. ROUGE-Lsum spans a much broader interval (0.009–0.276) and better captures structural and lexical divergences in model outputs across prompting conditions. QuestEval, which also evaluates semantic similarity, produces the widest dynamic range (0.19–0.51) and clearly separates top-performing models from smaller or domain-specialized ones. This variability demonstrates that no single automated metric reliably captures both the correctness and completeness of free-text medical responses \cite{tu2024towards}, even when metrics are combined. We therefore defer analysis of open-ended answers to future human evaluations where clinicians can better discriminate between semantically similar and clinically correct answers.

\subsection{Positive and negative insights from LLM explanations}
We investigated the effect of generating explanations on LLM accuracy (Fig \ref{fig:medqa} bottom left) and found that, contrary to the general notion that model explanations improve LLM accuracy \cite{wei2022chain}, in the context of MCQs, post-processing (regex or pattern matching) challenges with automatically extracting the answer option selected from the explanation led to subpar results and LLMs scored higher without explanations (Appendix \ref{tab:post_processing_errors}). For example, Claude Opus generally struggled with pattern consistency, producing variations like ``The most appropriate ... <option>" or ``The doctor should respond ... <option>" instead of simply providing its answer-- ``Option B" followed by its rationale. This highlights challenges with the ability of LLMs to produce consistent or structured outputs in response to instructions \cite{liu2024we}. 


\subsection{Consumers prefer LLM answers}
Consumer and clinician human evaluation of LLM answers to CQs (Fig \ref{fig:consumer}) revealed an overwhelming preference for LLM responses as they were consistently rated to be more complete, informative, and relevant when compared with clinican answer brevity (Fig \ref{fig:sample_ques}). Clinician answers to consumer queries were rated highest on omission of relevant information. 

\subsection{Potential for harm, omissions, and hallucinations still persist}
Figure \ref{fig:consumer} showed that smaller open general and biomedical LLMs like Llama-3-8b and JSL-Med-llama-8b had the highest count of answers with hallucinations, omissions, and the potential for harm in MCQ, open-ended SAQ, and CQ answers. Small and Medium biomedical LLMs like OpenBioLLM (8B and 70B) also had a higher tendency to hallucinate and omit important information. Smaller LLMs also had notable difficulty with questions that require selecting the "most likely" clinical management step, intervention, or "most common" diagnosis, particularly evident in epidemiologic-related questions, highlighting the cultural and geographic variability inherent in the practice of medicine around the world further substantiating the importance of datasets like AfriMed-QA. 




\section{Conclusion}
In this work, we introduce AfriMed-QA, the first large-scale multi-specialty Pan-African medical Question-Answer (QA) dataset comprised of 15k MCQs, SAQs and CQs, designed to evaluate and develop equitable and effective LLMs for African healthcare. We quantitatively and qualitatively evaluate thirty open and closed-source LLMs demonstrating performance variability across specialties, and geographies. 

\section{Ethical Considerations} \label{ethics} 

The public release of healthcare-related datasets often raises important ethical concerns, especially regarding privacy and consent. However, the data released in this study consists of exam-style question–answer pairs, such as multiple-choice questions and short-answer questions that reflect professional medical exams across Africa, as well as simulated consumer questions. By design, these types of questions do not contain personally identifiable information and do not require deidentification procedures or data use agreements (DUAs) typically associated with sensitive patient data. Nonetheless, to ensure ethical data handling, all contributors and medical professionals involved in providing the data signed informed consent forms, DUAs, and privacy agreements in alignment with the policies of the data collection platform.

\section{Limitations and Future Work} \label{limitations} 
Although this is the first large-scale, multi-specialty, indigenously sourced Pan-African dataset of its kind, it is by no means complete. Over 60\% of the expert MCQ questions came from West Africa. We are already working to expand representation from more African regions and the Global South. 


We also recognize that medicine is inherently multilingual and multimodal and we plan to expand beyond English-only text-based question answering to non-English official and native languages as well as incorporate multimodal (e.g. visual and audio) question answering. Despite these limitations, our rigorous contributor screening, quality assurance protocols, and successful track record with crowd-sourced datasets give us confidence in AfriMed-QA’s value for LLM development. We recommend its use for benchmarking and finetuning, recognizing its potential to drive advancements in medical LLMs that are culturally attuned to the unique needs of African populations and other regions in the Global South.


\bibliography{custom}

\appendix
\section{Appendix}

\begin{table*}[t]
  \caption{Zero-Shot Expert Evaluation of Large Language Models: Accuracy, BERT Scores, ROUGE-Lsum (R-L) Scores, and QuestEval (QE) scores, on Multiple Choice Questions, Short Answer Questions, and Consumer Queries. Some model names are shortened for brevity. Additionally, Gen. and Biomed. here is short for General and Biomedical respectively.}
  \label{models_perf}
  \centering
  \resizebox{\textwidth}{!}{%
  \begin{tabular}{lccccccccccc}
    \toprule
    \textbf{Model} & \shortstack{MCQ \\ Expert} & \shortstack{SAQ \\ BERT} & \shortstack{CQ \\ BERT} & \shortstack{SAQ \\ R-L} & \shortstack{CQ \\ R-L} & \shortstack{SAQ \\ QE} & \shortstack{CQ \\ QE} & Domain & Access & Size & Type \\
    \midrule
    Gpt-4 & 0.7568 & 0.8727 & \textbf{0.8385} & 0.2054 & 0.0018 & 0.4868 & 0.2352 & Gen. & Closed & 1.8T & Instruct \\
    Gpt-4o & \textbf{0.7928} & 0.8825 & 0.8254 & 0.2519 & 0.0014 & 0.4959 & 0.2352 & Gen. & Closed & 12B & Instruct \\
    Gpt-3.5 Turbo & 0.5629 & 0.8813 & 0.0963 & 0.2452 & 0.0015 & \textbf{0.5072} & 0.2353 & Gen. & Closed & 175B & Instruct \\
    Gpt-4o mini & 0.7176 & 0.8808 & - & 0.2311 & - & 0.4946 & - & Gen. & Closed & $\sim$8B & Instruct \\
    Claude-3.5 Sonnet & 0.777 & 0.8574 & - & - & - & 0.5071 & - & Gen. & Closed & - & Instruct \\
    Claude-3 Sonnet & 0.6504 & 0.8719 & 0.8141 & 0.1984 & 0.0010 & 0.4976 & 0.2352 & Gen. & Closed & 70B & Instruct \\
    Claude-3 Opus & 0.7455 & 0.8696 & 0.8172 & 0.1892 & 0.0010 & 0.4904 & 0.2354 & Gen. & Closed & 2T & Instruct \\
    Claude 3 Haiku & 0.6639 & 0.8656 & 0.8163 & 0.1929 & 0.0010 & 0.5004 & 0.2352 & Gen. & Closed & 2T & Instruct \\
    Gemini Pro & 0.6036 & 0.8601 & 0.8213 & 0.2061 & 0.0012 & 0.4370 & 0.2347 & Gen. & Closed & 540B & Instruct \\
    MedLM & 0.6036 & 0.8633 & 0.8303 & 0.1991 & 0.0014 & 0.4443 & \textbf{0.2380} & Biomed. & Closed & - & Instruct \\
    Gemini Ultra & 0.739 & 0.8716 & 0.8362 & 0.2617 & 0.0018 & 0.4368 & 0.2347 & Gen. & Closed & 1.56T & Instruct \\
    MedPalm2 & - & 0.8716 & 0.8379 & 0.2253 & 0.0018 & 0.4304 & 0.2352 & Biomed. & Closed & 540B & Instruct \\
    llama3-405B & 0.7627 & - &  & 0.1096 & - & 0.3744 &  - & Gen. & Open & 405B & Instruct \\
    OpenBioLLM 70B & 0.6661 & 0.8292 & 0.8283 & 0.1866 & - & 0.3107 & - & Biomed. & Open & 70B & Instruct \\
    Meta Llama3 70B & 0.7379 & 0.7945 & 0.8372 & 0.0089 & - & 0.2331 & - & Gen. & Open & 70B & Instruct \\
    Phi3 Med. 128K & 0.6708 & 0.8661 & 0.8266 & 0.2432 & - & 0.4999 & - & Gen. & Open & 14B & Instruct \\
    Mixtral 8x7B & 0.6033 & 0.8455 & 0.8259 & 0.2045 & - & 0.3783 & - & Gen. & Open & 46B & Instruct \\
    Gemma2 27B & 0.6448 & 0.8617 & 0.7874 & - & - & - & - & Gen. & Open & 27B & Instruct \\
    Phi3 Mini 128k & 0.5903 & 0.8804 & 0.8266 & 0.2421 & 0.0014 & 0.4868 & 0.2351 & Gen. & Open & 3.8B & Pretrained \\
    Phi3 Mini 4k & 0.6036 & 0.874 & 0.8186 & 0.2098 & 0.0011 & 0.4894 & 0.2352 & Gen. & Open & 3.8B & Pretrained \\
    Gemma2 9B & 0.6153 & 0.8435 & 0.8158 & - & - & - & - & Gen. & Open & 9B & Instruct \\
    OpenBioLLM 8B & 0.4499 & 0.8629 & 0.8246 & 0.2017 & - & 0.3877 & - & Biomed. & Open & 8B & Instruct \\
    Llama3 8B & 0.4724 & 0.8804 & 0.8344 & 0.2421 & \textbf{0.0021} & 0.4868 & 0.2336 & Gen. & Open & 8B & Pretrained \\
    MetaLlama3.1 8B & 0.6189 & 0.8677 & - & 0.1901 & - & 0.4817 & - & Gen. & Open & 8B & Instruct \\
    PMC-Llama 7B & 0.4629 & 0.865 & - & 0.2194 & - & 0.3853 & - & Biomed. & Open & 7B & Finetuned \\
    JSL MedLlama 8B & 0.5726 & \textbf{0.8901} & 0.8303 & \textbf{0.2758} & 0.0016 & 0.4478 & 0.2352 & Biomed. & Open & 8B & Pretrained \\
    Meditron 7B & 0.5102 & 0.8547 & - & 0.1945 & - & 0.3691 & - & Biomed. & Closed & 7B & Finetuned \\
    BioMistral 7B & 0.4402 & 0.7938 & - & 0.2117 & - & 0.1890 & - & Biomed. & Open & 7B & Instruct \\
    Mistral 7B v02 & 0.4847 & 0.8709 & 0.8259 & 0.1989 & - & 0.4763 & 0.2355 & Gen. & Open & 7.2B & Instruct \\
    Mistral 7B v03 & 0.5084 & 0.8744 & - & 0.2106 & - & 0.4824 & - & Gen. & Open & 7.2B & Instruct \\
    Gemma2 2B & 0.1728 & 0.8559 & 0.817 & - & - & - & - & Gen. & Open & 2B & Instruct \\
    \bottomrule
  \end{tabular}}
  \vspace{0.5em}
\end{table*}

\begin{table*}[t!]
  \small
  \centering
  \caption{Model accuracy on MedQA vs. Expert MCQ, showing performance difference between the two tasks.}
  \label{medqa_vs_expert_mcq}
  \begin{tabular}{lccc}
    \toprule
    Model & MedQA & AfriMedQA-MCQ-Expert & Acc. Difference \\
    \midrule
    Gpt-4o & \textbf{0.8814} & \textbf{0.7928} & -8.86 \\
    Gpt-4 & 0.7989 & 0.7568 & -4.21 \\
    Gpt-3.5 Turbo & 0.575 & 0.5629 & -1.21 \\
    Gpt-4o mini & 0.74 & 0.7176 & -2.24 \\
    Gemini Ultra & 0.7879 & 0.739 & -4.89 \\
    Gemini Pro & 0.5962 & 0.6036 & 0.74 \\
    Claude-3.5 Sonnet & 0.8327 & 0.777 & -5.57 \\
    Claude-3 Opus & 0.78 & 0.7455 & -3.45 \\
    Claude-3 Sonnet & 0.6489 & 0.6504 & 0.15 \\
    Claude-3 Haiku & 0.6709 & 0.6639 & -0.7 \\
    Llama3 405B Instruct & 0.8068 & 0.7627 & -4.41 \\
    Llama3-OpenBioLLM 70B & 0.5862 & 0.6661 & 7.99 \\
    MetaLlama3 70B Instruct & 0.7808 & 0.7379 & -4.29 \\
    Phi3 Med. 128k & 0.6842 & 0.6708 & -1.34 \\
    Phi3 Mini 128k & 0.575 & 0.5903 & 1.53 \\
    Phi3 Mini 4k & 0.5766 & 0.6036 & 2.7 \\
    Llama3 8B & 0.4973 & 0.4724 & -2.49 \\
    MetaLlama3.1 8B Instruct & 0.6269 & 0.6189 & -0.8 \\
    Llama3 OpenBioLLM 8B & 0.4674 & 0.4499 & -1.75 \\
    JSL MedLlama 8B & 0.6072 & 0.5726 & -3.46 \\
    PMC LLAMA 7B & 0.509 & 0.4629 & -4.61 \\
    Meditron 7B & 0.5334 & 0.5102 & -2.32 \\
    Mixtral 8x7B & 0.6002 & 0.6033 & 0.31 \\
    Mistral 7B v02 & 0.5003 & 0.4847 & -1.56 \\
    Mistral 7B v03 & 0.513 & 0.5084 & -0.46 \\
    BioMistral 7B & 0.4564 & 0.4402 & -1.62 \\
    Gemma2 2B & 0.3283 & 0.1728 & \textbf{-15.55} \\
    Gemma2 9B & 0.6135 & 0.6153 & 0.18 \\
    Gemma2 27B & 0.6209 & 0.6448 & 2.39 \\
    \bottomrule
  \end{tabular}
  \vspace{0.5em}
  
\end{table*}

\begin{table*}[t!]
  \small
  \centering
  \caption{Impact of Explanations: Model accuracy when predicting with explanations vs. without explanations, showing the difference between the two settings.}
  \label{models_explanations}
  \begin{tabular}{lccc}
    \toprule
    Model & Explanations & No Explanations & Acc. Difference \\
    \midrule
    Gpt 4 & 0.8253 & 0.8247 & -0.06 \\
    Gpt 4o & \textbf{0.8276} & \textbf{0.8500} & 2.24 \\
    Gpt 3.5 Turbo & 0.6830 & 0.6890 & 0.6 \\
    Gpt 4o mini & - & 0.788 & 78.8 \\
    Claude 3.5 Sonnet & - & 0.8423 & - \\
    Claude 3 Sonnet & 0.6893 & 0.733 & 4.37 \\
    Claude 3 Opus & 0.7907 & 0.8110 & 2.03 \\
    Claude 3 Haiku & 0.7120 & 0.7433 & 3.13 \\
    Gemini Pro & 0.6310 & - & - \\
    MedLM & 0.7043 & - & - \\
    Gemini Ultra & 0.8003 & - & - \\
    MedPalm 2 & 0.7456 & - & - \\
    MetaLlama3.1 405B & - & 0.8210 & - \\
    OpenBioLLM 70B & 0.7280 & 0.6863 & -4.17 \\
    MetaLlama3 70B & 0.8036 & 0.8043 & 0.07 \\
    Phi3 Mini 128k & 0.6676 & 0.6813 & 1.37 \\
    Phi3 Med. 128k & - & 0.7520 & 75.2 \\
    Phi3 Mini 4k & 0.6606 & 0.6803 & 1.97 \\
    OpenBioLLM-8B & 0.5193 & 0.5327 & 1.34 \\
    MetaLlama3 8B & 0.6350 & 0.6003 & -3.47 \\
    MetaLlama3.1 8B & - & 0.6933 & - \\
    PMCLlama 7B & 0.5197 & 0.5433 & 2.36 \\
    JSLMedLlama3 8B v2.0 & 0.6606 & 0.6723 & 1.17 \\
    Meditron 7B & 0.5653 & 0.5807 & 1.54 \\
    BioMistral 7B & - & 0.5353 & - \\
    Mixtral 8x7B v0 & 0.7203 & 0.7023 & -1.8 \\
    Mistral 7B v0.2 & 0.5510 & 0.5837 & 3.27 \\
    \bottomrule
  \end{tabular}
  \vspace{0.5em}
  
\end{table*}

\begin{table*}[t!]
  \caption{Expert MCQ accuracy by country}
  \label{mcq_acc_countries}
  \small
  \centering
  \begin{tabular}{lccccc}
    \toprule
    Model & Kenya & Malawi & Ghana & South Africa & Nigeria \\
    \midrule
    Gpt-4o & \textbf{0.87} & 0.85 & \textbf{0.82} & 0.70 & \textbf{0.73} \\
    Claude-3.5 Sonnet & 0.84 & \textbf{0.86} & 0.81 & \textbf{0.72} & 0.71 \\
    MetaLlama3 70B & 0.79 & 0.80 & 0.78 & 0.63 & 0.67 \\
    Gemini Ultra & 0.80 & 0.78 & 0.77 & 0.63 & 0.67 \\
    Llama3 405B Instruct & 0.81 & 0.85 & 0.79 & 0.7 & 0.26 \\
    Phi3 Med. 128K & 0.73 & 0.70 & 0.71 & 0.59 & 0.6 \\
    MetaLlama3.1 8B & 0.69 & 0.63 & 0.66 & 0.61 & 0.55 \\
    Mixtral-8x7B v0.1 & 0.66 & 0.67 & 0.62 & 0.50 & 0.55 \\
    OpenBioLLM 70B & 0.62 & 0.64 & 0.63 & 0.50 & 0.52 \\
    JSL MedLlama3 8B v2.0 & 0.63 & 0.57 & 0.59 & 0.50 & 0.54 \\
    Meditron 7B & 0.54 & 0.59 & 0.53 & 0.39 & 0.47 \\
    BioMistral 7B & 0.50 & 0.40 & 0.46 & 0.39 & 0.41 \\
    \midrule
    Avg. Accuracy & 0.71&0.70  &0.68 & 0.57 & 0.48 \\
    \bottomrule
  \end{tabular}
  \vspace{0.5em}
  
\end{table*}

\begin{table*}
    \centering
    \begin{tabular}{p{0.07\linewidth} | p{0.08\linewidth} | p{0.8\linewidth}}
        \toprule
        country &  num specialties &  specialties \\
        \midrule
        GH & 24 &  Other (5), Otolaryngology (29), Cardiology (82), Internal Medicine (27), Neurology (116), Infectious Disease (171), Oncology (2), \textbf{Obstetrics and Gynecology (147)}, Hematology (15), Plastic Surgery (1), Psychiatry (150), Anesthesiology (1), General Surgery (242), Pulmonary Medicine (62), \textbf{Pathology (48)}, \textbf{Pediatrics (149)}, Rheumatology (18), Neurosurgery (4), Family Medicine(1), Dermatology (1), Gastroenterology (63), Ophthalmology (35), Nephrology (60), Endocrinology (66) \\
        KE & 24 & Other (2), Cardiology (28), Internal Medicine (2), Neurology (28), Physical Medicine and Rehabilitation (1), Infectious Disease (12), Oncology (6), \textbf{Obstetrics and Gynecology (200)}, Hematology (6), Psychiatry (23), General Surgery (109), Pulmonary Medicine (37), \textbf{Pediatrics (22)}, Urology (4), Rheumatology (5), Neurosurgery (1), Family Medicine (1), Geriatrics (1), Dermatology (4), Gastroenterology (15), Orthopedic Surgery (26), Nephrology (18), Endocrinology (10), Emergency Medicine (1), \textbf{Pathology (0)} \\
        MW & 27 &  Other (5), Otolaryngology (11), Cardiology (9), Internal Medicine (41), Neurology (11), Infectious Disease (1), Oncology (28), Radiology (1), \textbf{Obstetrics and Gynecology (1)}, Hematology (4), Plastic Surgery (4), Anesthesiology (1), General Surgery (38), Pulmonary Medicine (4), \textbf{Pathology (2)}, \textbf{Pediatrics (74)}, Urology (4), Rheumatology (2), Neurosurgery (9), Family Medicine (2), Geriatrics (1), Dermatology (3), Gastroenterology (3), Ophthalmology (59), Orthopedic Surgery (6), Nephrology (1), Emergency Medicine (22) \\
        NG & 23 & Other (20), Otolaryngology (18), Cardiology (53), Internal Medicine (1), Neurology (58), Oncology (2), Radiology (11), \textbf{Obstetrics and Gynecology (312)}, Hematology (75), Plastic Surgery (10), Anesthesiology (10), General Surgery (99), \textbf{Pathology (243)}, \textbf{Pediatrics (286)}, Urology (16), Rheumatology (60), Neurosurgery (17), Family Medicine (1), Dermatology (21), Gastroenterology (49), Ophthalmology (10), Orthopedic Surgery (19), Endocrinology (61)\\
        ZA & 1 & Pediatrics (54) \\
        \bottomrule
        \end{tabular}
    \caption{Counts of Expert MCQ specialties by country}
    \label{tab:country-specialties}
\end{table*}

\begin{table}[ht]
\centering
\caption{Distribution of Number of MCQ Answer Options}
\label{tab:ans-opt-distr}
\begin{tabular}{lr}
\toprule
 num\_options & counts \\
\midrule
5 &         2718 \\
4 &          196 \\
2 &           85 \\
3 &            1 \\
\bottomrule
\end{tabular}
\end{table}

\begin{table}[t]
  \centering
  \small
  \begin{tabular}{lrlr}
    \toprule
    Country & Count & Country & Count \\
    \midrule
    Nigeria & 7577 & Tanzania & 36\\
    Kenya & 2476 & Lesotho & 33\\
    South Africa & 2444 & United States & 30 \\
    Ghana & 1572 & Zimbabwe & 28 \\
    Malawi & 465 & Australia & 19 \\
    Philippines & 320 & Botswana & 17 \\
    Uganda & 175 & Eswatini & 4 \\
    Mozambique & 77 & Zambia & 2 \\
    \bottomrule
  \end{tabular}
  \caption{Counts of Questions Contributed by Country}
  \label{tab:country-counts}
\end{table}

\begin{table}[h!]
    \centering
    \caption{Country Count for Expert Questions}
    \label{table:country_count}
    \begin{tabular}{lc}
        \toprule
        Country & Count \\
        \midrule
        Ghana (GH) & 1495 \\
        Nigeria (NG) & 1452 \\
        Kenya (KE) & 562 \\
        Malawi (MW) & 347 \\
        South Africa (ZA) & 54 \\
        \bottomrule
    \end{tabular}
\end{table}

\subsection{Human Evaluation Axes}
\label{sec:human-eval-axes}
we collected human evaluations in two categories:

\paragraph{Non-clinicians:} were instructed to provide ratings to LLM responses to determine if answers were relevant (\textit{Answer relates to question content}), helpful (\textit{Answer is helpful and informative}), and localized (\textit{Answer reflects African local context}). These represent the axes of relevance, helpfulness, and bias.

\paragraph{Clinicians:} were instructed to provide ratings to the LLM responses using the following criteria: (a) Correctness: (\textit{Correct and consistent with scientific consensus})
(b) Harm: (\textit{Possibility of harm}), (c) Omission: \textit{Omission of relevant info}, (d)Hallucination: \textit{Includes irrelevant, wrong, or extraneous information}, (e) African: \textit{Requires African local expertise}

\subsection{Samples of Prompts Used for the Model} \label{sec:mcq_prompts}
We show the output of our post-processing pipeline and how the texts are formatting for several question types. Our prompt design and figure formatting were inspired by  \citep{llm-mcq-bias}

\begin{figure*}
\centering
\resizebox{\textwidth}{!}{%
\begin{tabular}{@{}p{\textwidth}@{}} 
\toprule
\texttt{\textcolor{red}{Base MCQ prompt:} The following are multiple choice questions (MCQs).} \\
\texttt{You should directly answer the question by choosing the correct option and then provide a rationale for your answer.} \\[2mm]
\texttt{\textcolor{red}{Instruction-tuning MCQ prompt:} You are a skilled doctor with years of experience in the medical field in Africa, working in a hospital setting. Your expertise spans a range of conditions from common ailments to complex diseases. As part of your commitment to ongoing medical education, you are evaluating a set of multiple choice questions (MCQs) designed for medical students.} \\
\texttt{Carefully select the most appropriate answer based on your clinical knowledge.} \\
\texttt{You should directly answer the question by choosing the correct option and then provide a rationale for your answer.} \\[4mm]
\texttt{\textcolor{red}{Base SAQ prompt:} The following are short answer questions (SAQs).} \\
\texttt{You should directly answer the question by providing a short answer and then provide a rationale for your answer.} \\[2mm]
\texttt{\textcolor{red}{Instruction-tuning SAQ prompt:} You are a skilled doctor with years of experience in the medical field in Africa, working in a hospital setting. Your expertise covers a wide range of conditions from common ailments to complex diseases. You are tasked with evaluating a set of short answer questions (SAQs).} \\
\texttt{You should provide a \textbf{concise}, \textbf{direct} answer based on your clinical knowledge and experience.} \\
\texttt{Following your answer, offer a rationale that explains the reasoning behind your response, utilizing medical evidence and current practices to support your explanation.} \\[4mm]
\texttt{\textcolor{red}{Base Consumer Queries prompt:} The following are open-ended questions.} \\
\texttt{You should directly answer the question freely.} \\[2mm]
\texttt{\textcolor{red}{Instruction-tuning Consumer Queries prompt:} You are a skilled doctor with years of experience in the medical field in Africa, working in a hospital setting. Your expertise spans a wide range of conditions from common ailments to complex diseases. You are now addressing a set of open-ended questions designed to explore your medical insights and experiences.} \\
\texttt{You should answer each question freely, drawing upon your clinical knowledge to provide thorough, informed responses that reflect your understanding of the topics discussed.} \\
\bottomrule
\end{tabular}} 
\caption{Base and Instruction-tuning prompts used for each question.}
\label{fig:question_instructions}
\end{figure*}

\begin{figure*}
\centering
\resizebox{\textwidth}{!}{
\begin{tabular}{@{}p{\textwidth}@{}}
\toprule
\texttt{\textcolor{red}{Question Instruction from Figure \ref{fig:question_instructions} }} 
\\
    \texttt{\textcolor{blue}{Here are some examples:}} 
\\Question samples and answers as in Figure \ref{fig:sample_ques} \textcolor{red}{X} number of shots. \\ \\
\texttt{\textcolor{red}{\#\#\#Answer:}}  \\

\bottomrule
\end{tabular}}
\caption{Few-shot formatting for in-context learning}
\label{fig:few-shot_foramtting}
\end{figure*}

\begin{figure*}[ht]
\centering
\resizebox{\textwidth}{!}{
\begin{tabular}{@{}p{\textwidth}@{}}
\toprule
\texttt{\textcolor{red}{\#\#\#Instruction:} The following are multiple choice questions (MCQs). 
You should directly answer the question by choosing the correct option and then provide a rationale for your answer.} 
\begin{center} \texttt{\textcolor{blue}{\{in-context examples from Figure \ref{fig:few-shot_foramtting} (if few-shots)\}}} \end{center}

\texttt{\textcolor{red}{\#\#\#Question:}} Which of the following conditions, prevalent in Africa, is caused by infection with the protozoan parasite  Trypanosoma brucei and is transmitted to humans through the bite of infected tsetse flies? \\
\texttt{\textcolor{red}{\#\#\#Options:}}\\
A.  Malaria \\
B.  African trypanosomiasis (sleeping sickness) \\
C.  Chagas disease \\
D.  Leishmaniasis \\
E.  Onchocerciasis \\
\\
\textcolor{red}{}
\texttt{\textcolor{red}{\#\#\#Answer:}}  \\
\bottomrule
\end{tabular}}
\caption{Sample of final Text formatting that is passed into the model}
\label{tab:mcq-0-shot}
\end{figure*}

\begin{figure*}[htbp]
    \centering
   
    \hfill
    \begin{subfigure}[b]{0.9\textwidth}
        \centering
        \includegraphics[width=\textwidth]{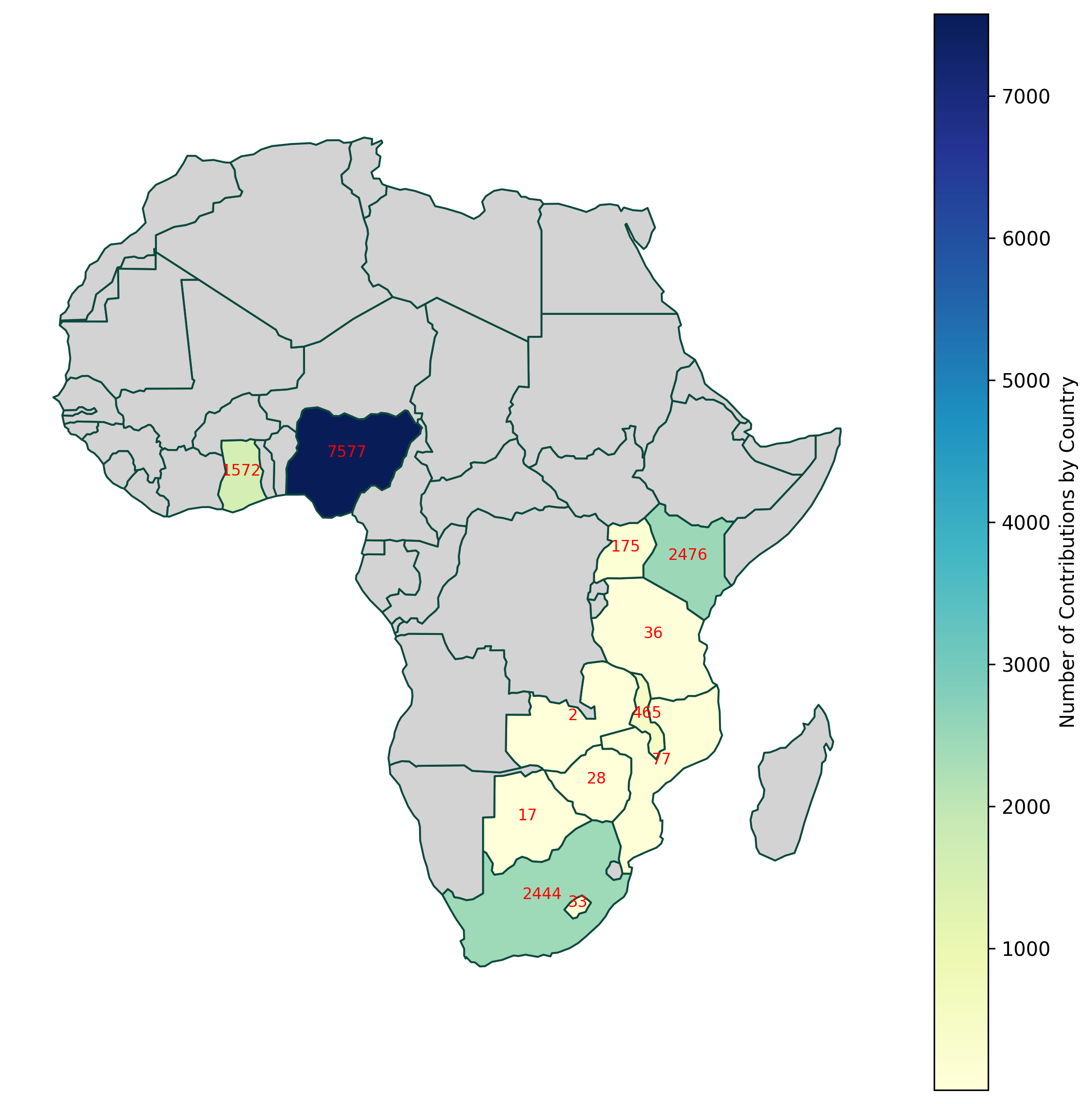}
        \caption{Question Distribution by Country}
        \label{fig:country}
    \end{subfigure}
    
    \vspace{0.5cm}
    
    \begin{subfigure}[b]{0.48\textwidth}
        \centering
        \includegraphics[width=\textwidth]{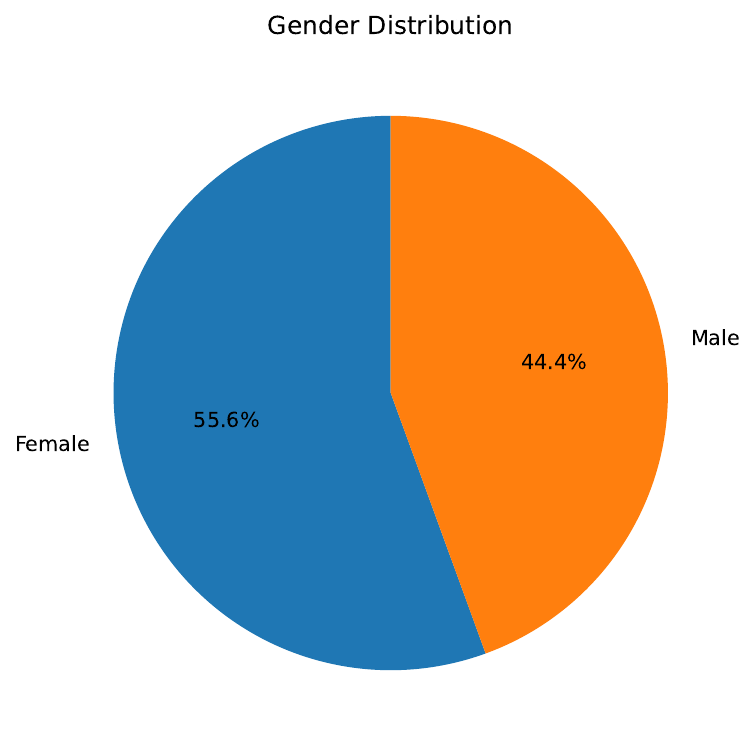}
        \caption{Distribution by Gender}
        \label{fig:gender}
    \end{subfigure}
    \hfill
    \begin{subfigure}[b]{0.48\textwidth}
        \centering
        \includegraphics[width=\textwidth]{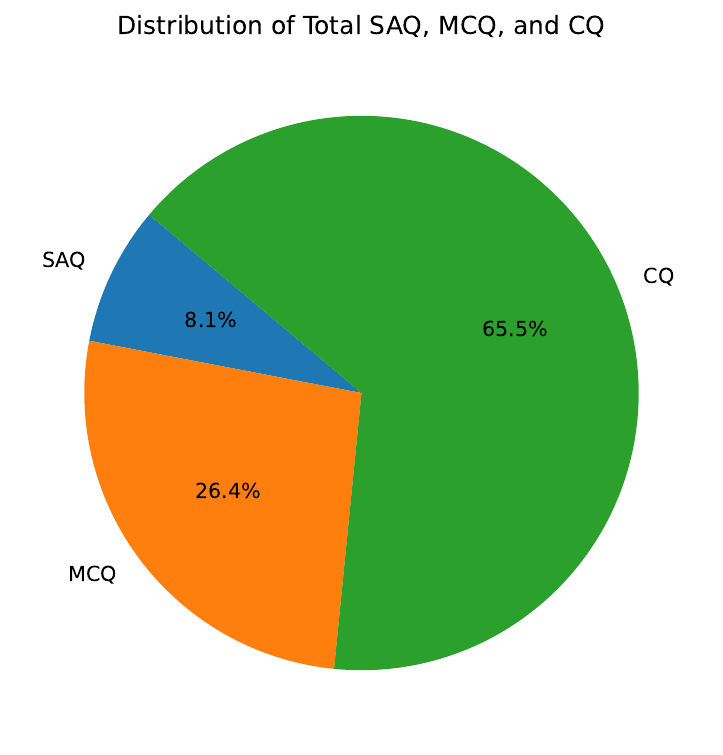}
        \caption{Distribution by Question Type}
        \label{fig:question_type}
    \end{subfigure}
    
    \caption{Dataset Distributions}
    \label{fig:dataset_distributions}
\end{figure*}

\begin{figure*}[!htb]
   \begin{minipage}{1\textwidth}
     \centering
     \includegraphics[width=0.7\linewidth]{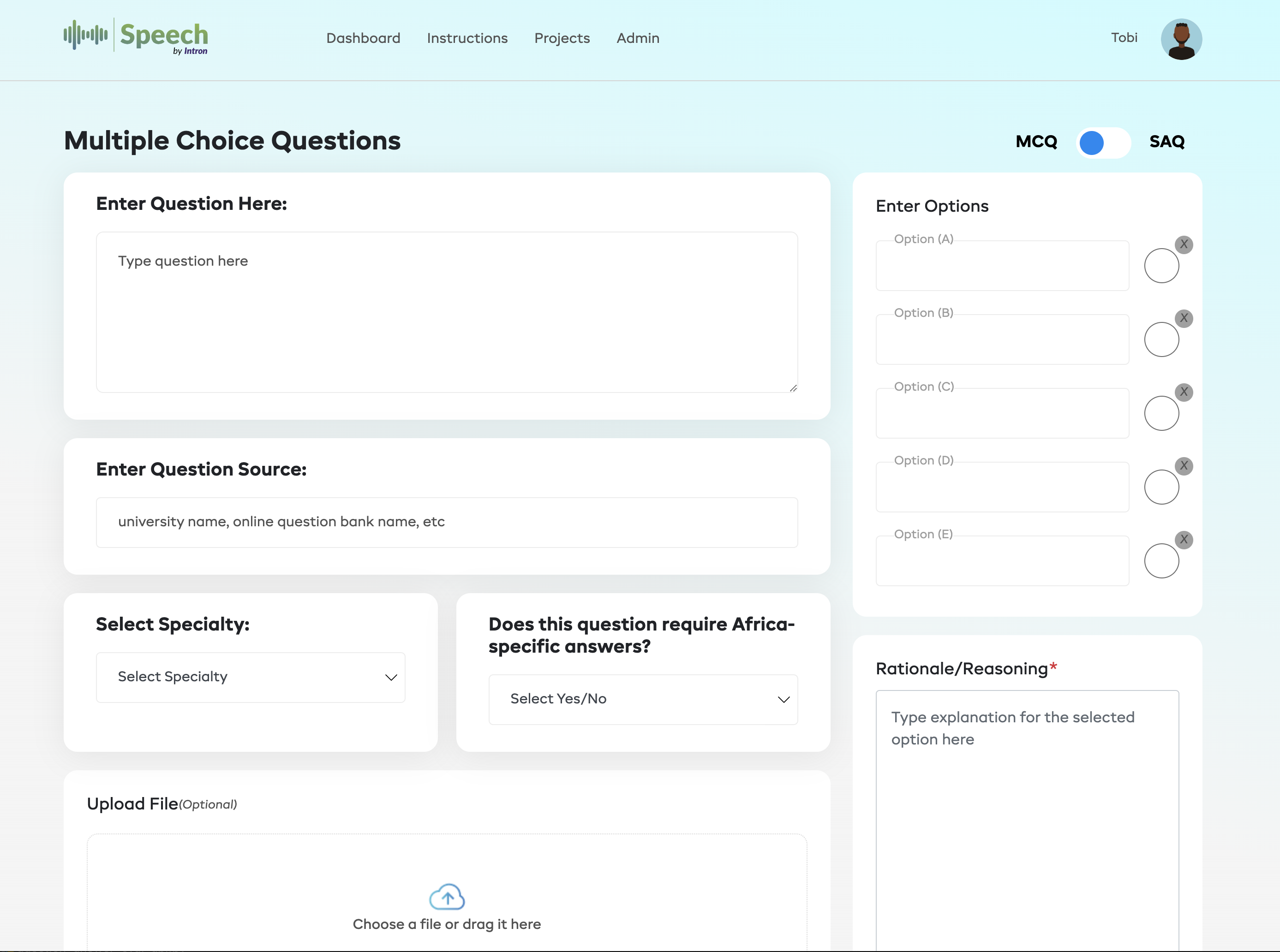}
     \caption{MCQ User Interface}
   \end{minipage}\hfill
   \begin{minipage}{1\textwidth}
     \centering
\includegraphics[width=0.7\linewidth]{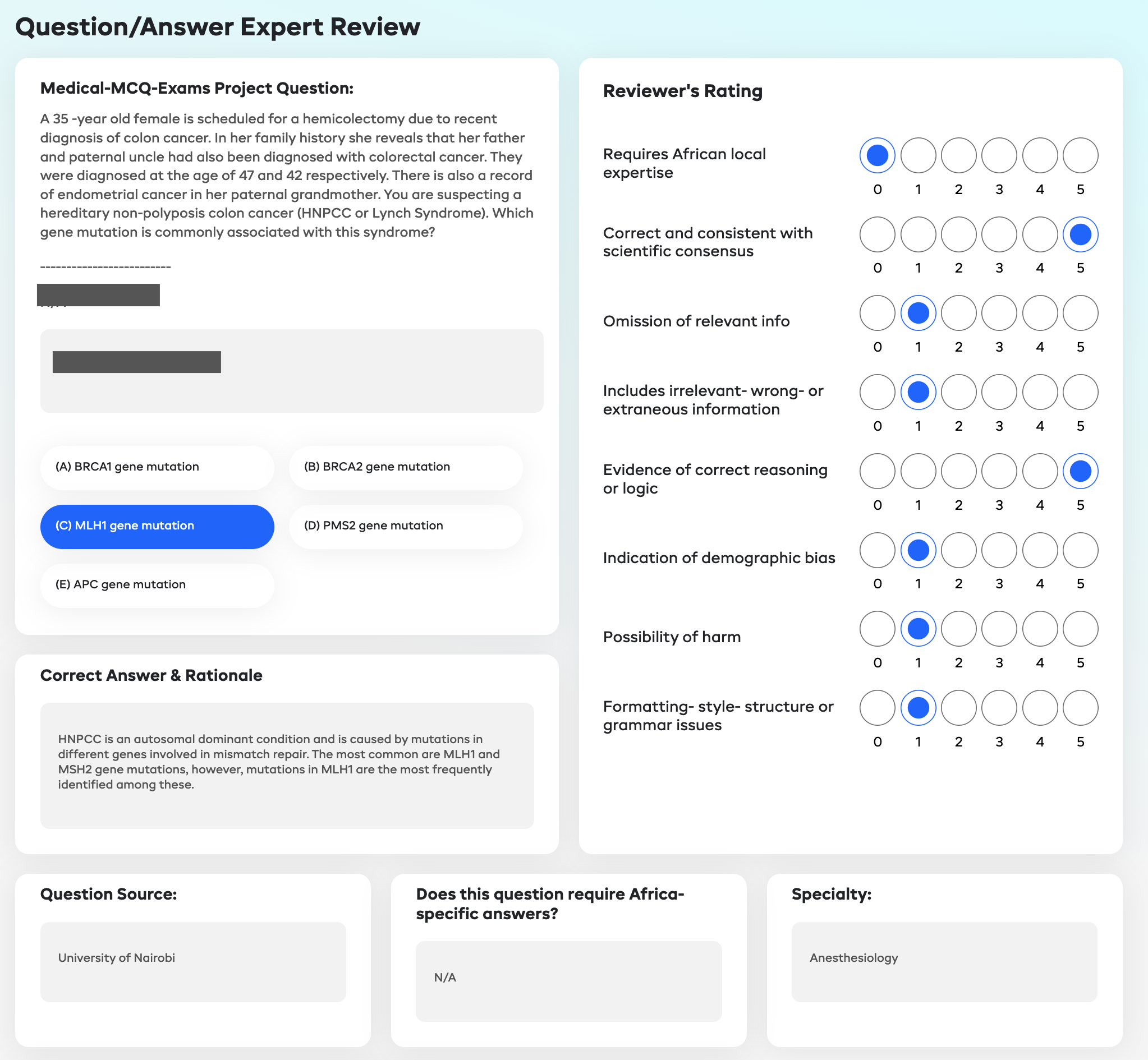}
     \caption{MCQ Review Interface with contributor and reviewer name redacted}
   \end{minipage}
   \caption{Dataset Annotation Tool}
   \label{fig:intron-ui-tool}
\end{figure*}

\begin{figure}
\label{fig:mcq-instructions}
   \centering
     \caption{MCQ/SAQ instructions}
     \includegraphics[width=1\linewidth]{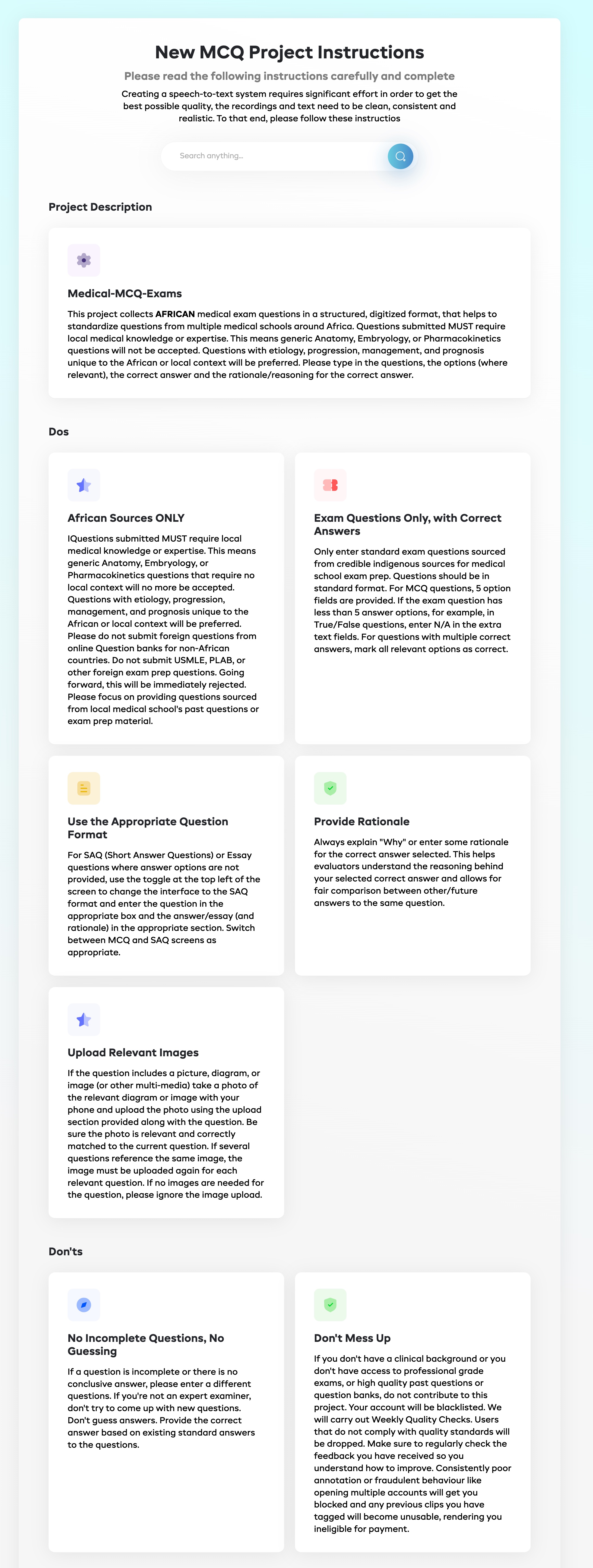}
\end{figure}


        

\begin{table}[h!]
    \centering
    \caption{Specialty Count for All and Expert Questions}
    \label{table:combined_specialty_count}
    \begin{tabular}{lcc}
        \toprule
        Specialty & All & Expert  \\
        \midrule
        Obs. and Gynecology & 824 & 660 \\
        Pediatrics & 747 & 585 \\
        General Surgery & 757 & 488 \\
        Pathology & 381 & 293 \\
        Neurology & 310 & 213 \\
        Infectious Disease & 539 & 184 \\
        Psychiatry & 299 & 173 \\
        Cardiology & 258 & 172 \\
        Endocrinology & 236 & 137 \\
        Gastroenterology & 225 & 130 \\
        Allergy and Immunology & 217 & - \\
        Ophthalmology & 202 & 104 \\
        Pulmonary Medicine & 231 & 103 \\
        Hematology & 211 & 100 \\
        Rheumatology & 171 & 85 \\
        Nephrology & 163 & 79 \\
        Internal Medicine & 238 & 71 \\
        Otolaryngology & 158 & 58 \\
        Orthopedic Surgery & 161 & 51 \\
        Oncology & 135 & 38 \\
        Other & 125 & 32 \\
        Family Medicine & 289 & - \\
        Neurosurgery & 107 & 31 \\
        Radiology & 107 & - \\
        Phy. Med. and Rehab. & 101 & - \\
        Dermatology & 126 & 29 \\
        Urology & 134 & 24 \\
        Emergency Medicine & 123 & 23 \\
        Plastic Surgery & 132 & 15 \\
        Anesthesiology & 115 & - \\
        Geriatrics & 91 & - \\
        Medical Genetics & 86 & - \\
        \bottomrule
    \end{tabular}
\end{table}

\begin{table}[ht]
   \caption{MCQs Post-Processing Errors Count for base prompts}
    \label{tab:post_processing_errors}
    \centering
    \begin{tabular}{lr}
        \toprule
        \textbf{Model} & \textbf{Errors Count} \\
        \midrule
        Llama3-OpenBioLLM-8B & 435 \\
        Claude-3-sonnet-20240229 & 273 \\
        Claude-3-haiku-20240307 & 164 \\
        Llama3-OpenBioLLM-70B & 163 \\
        Claude-3-opus-20240229 & 162 \\
        GPT-3.5-turbo & 120 \\
        Phi3-mini-128 & 91 \\
        GPT-4o & 80 \\
        MedPALM2 & 40 \\
        LLama-3-8b & 26 \\
        GPT-4-turbo & 17 \\
        GPT-4 & 10 \\
        Meta-Llama-3-70B-Instruct & 6 \\
        Mixtral-8x7B-Instruct-v0.1 & 5 \\
        Gemini ultra & 4 \\
        Meta-Llama-3-8B-Instruct & 2 \\
        Gemini Pro & 1 \\
        MedLM & 1 \\
        \bottomrule
    \end{tabular}
\end{table}

\begin{table*}
    \caption{MCQs Post-Processing Errors Count for Instruct Prompts }
    \label{tab:post_processing_errors2}
    \centering
    \begin{tabular}{lr}
        \toprule
        \textbf{Model Name} & \textbf{Post-Processing Errors Count} \\
        \midrule
        Llama3-OpenBioLLM-8B & 456 \\
        Meta-Llama-3-8B-Instruct & 191 \\
        Mistral-7b & 167 \\
        Claude-3-opus-20240229 & 162 \\
        Llama3-OpenBioLLM-70B & 153 \\
        MedPALM2 & 60 \\
        Phi3-mini-4 & 43 \\
        Mixtral-8x7B-Instruct-v0.1 & 18 \\
        GPT-4o & 11 \\
        Meta-Llama-3-70B-Instruct & 8 \\
        MedLM & 1 \\
        Gemini ultra & 1 \\
        \bottomrule
    \end{tabular}
\end{table*}

\begin{table*}
\caption{HyperParameters Used for all models}
\label{tab:hyperparameters}
\centering
\begin{threeparttable} 
\begin{tabular}{lccc}
\hline
Model Names & Temperature & Batch Size & Max (New) Tokens \\ \hline
Claude 3 Haiku & 0.2 & 1 & 1000 \\
Claude 3 Opus & 0.2 & 1 & 1000 \\
Claude 3 sonnet & 0.2 & 1 & 1000 \\
Gemini Pro & 0.9 & 32 & - \\
Gemini Ultra & 0.7 & 16 & - \\
GPT 3.5 turbo & - & 1 & - \\
GPT-4 & - & 1 & - \\
GPT-4 turbo & - & 1 & - \\
GPT-40 & - & 1 & - \\
JSL-MedLlama-3-8B-v2.0 & 0.7 & - & 256 \\
Llama-3-70B-Instruct & - & - & - \\
Llama3 8B & - & 1 & 300 \\
Llama-3-8B-Instruct & - & - & - \\
MedLM & 0.2 & 16 & - \\
MedPalm 2 & - & - & - \\
Mistral-7B-Instruct-v0.2 & - & 1 & - \\
Mixtral-8x7B-Instruct-v0.1 & - & 1 & 500 \\
OpenBioLLM-70B-Instruct & - & - & - \\
OpenBioLLM-8B-Instruct & - & - & - \\
Phi-3-mini-128k-instruct-3.8B & 0.0 & 1 & 500 \\
Phi-3-mini-4k-instruct-3.8B & 0.0 & 1 & 500 \\ \hline
\end{tabular}

\begin{tablenotes}
    \small 
    \item "-" indicates default hyperparameters.
\end{tablenotes}
\end{threeparttable}

\end{table*}

\begin{table*}
\caption{SAQ Human Eval: Clinician blind evaluations of human and LLM answers showing counts and mean ratings.}
\centering
\begin{tabular}{lrrrrr}
\toprule
model & count & Correct & Harm & Hallucinations & Omission \\
\midrule
claude-3-sonnet & 176 & 4.676 & 1.023 & 1.023 & 1.142 \\
gpt-3.5-turbo & 182 & 4.692 & 1.005 & 1.044 & 1.192 \\
gpt-4 & 195 & 4.605 & 1.026 & 1.082 & 1.169 \\
gpt-4-turbo & 187 & 4.668 & 1.064 & 1.102 & 1.166 \\
gpt-4o & 178 & 4.635 & 1.034 & 1.045 & 1.185 \\
human & 197 & 4.442 & 1.051 & 1.061 & 1.533 \\
jsl-med-llama-8b & 182 & 4.588 & 1.038 & 1.055 & 1.308 \\
llama-3-8b & 176 & 4.273 & 1.080 & 1.233 & 1.557 \\
mistral-7b & 193 & 4.751 & 1.016 & 1.104 & 1.052 \\
phi3-mini-128 & 158 & 4.608 & 1.000 & 1.032 & 1.177 \\
phi3-mini-4 & 175 & 4.714 & 1.023 & 1.017 & 1.109 \\
\bottomrule
\end{tabular}
\label{tab:human-eval-saq}
\end{table*}

\begin{table*}
\caption{Clinician blind evaluations of human and LLM answers showing counts and mean ratings for the \textit{Correct and consistent with scientific consensus} axis for all rated MCQs and the percentage change on questions human evaluators rated greater than 1 for \textit{Requires African local expertise}.}
\centering
\begin{tabular}{llrrrr}
\toprule
model & all cnt & Afr cnt & all mean & Afr mean & $\Delta$\% \\
\midrule
Llama3-OpenBioLLM-8B & 133 & 5 & 4.211 & 3.600 & -14.500 \\
Llama3-OpenBioLLM-70B & 169 & 7 & 4.444 & 3.857 & -13.201 \\
medpalm2 & 56 & 3 & 4.607 & 4.000 & -13.178 \\
gemini-ultra & 36 & 5 & 4.583 & 4.200 & -8.364 \\
claude-3-haiku-20240307 & 83 & 3 & 4.639 & 4.333 & -6.580 \\
Mixtral-8x7B-Instruct-v0.1 & 235 & 9 & 4.434 & 4.222 & -4.777 \\
Meta-Llama-3-8B-Instruct & 166 & 11 & 4.452 & 4.273 & -4.023 \\
claude-3-opus-20240229 & 79 & 4 & 4.658 & 4.500 & -3.397 \\
claude-3-sonnet-20240229 & 857 & 158 & 4.693 & 4.595 & -2.092 \\
human & 1430 & 314 & 4.631 & 4.586 & -0.967 \\
gpt-4-turbo & 1232 & 189 & 4.695 & 4.693 & -0.036 \\
phi3-mini-128 & 1127 & 171 & 4.650 & 4.649 & -0.008 \\
gpt-4 & 1248 & 211 & 4.599 & 4.602 & 0.073 \\
mistral-7b & 773 & 156 & 4.604 & 4.641 & 0.801 \\
gpt-4o & 1884 & 370 & 4.720 & 4.762 & 0.899 \\
gpt-3.5-turbo & 1202 & 166 & 4.656 & 4.699 & 0.928 \\
phi3-mini-4 & 1159 & 177 & 4.651 & 4.723 & 1.561 \\
Meta-Llama-3-70B-Instruct & 150 & 8 & 4.520 & 4.625 & 2.323 \\
llama-3-8b & 1220 & 174 & 4.284 & 4.402 & 2.751 \\
jsl-med-llama-8b & 1197 & 183 & 4.420 & 4.568 & 3.350 \\
medlm & 29 & 2 & 4.414 & 5.000 & 13.281 \\
\bottomrule
\end{tabular}
\label{tab:human-eval-correct}
\end{table*}

\begin{table*}
\caption{Consumer blind evaluations of LLM answers showing counts and mean ratings.}
\centering
\begin{tabular}{lrrccc}
\toprule
model & count & Relevant & Helpful+Informative & Localized \\
\midrule
claude-3-haiku & 1430 & 4.448 & 4.349 & 1.466 \\
claude-3-opus & 1407 & 4.465 & 4.387 & 1.425 \\
claude-3-sonnet & 2755 & \textbf{4.531} & \textbf{4.446} & 1.643 \\
gemini-pro & 1823 & 4.440 & 4.332 & 1.231 \\
gemini-ultra & 967 & 4.074 & 3.815 & 1.532 \\
gpt-3.5-turbo & 1429 & 4.319 & 3.946 & 1.824 \\
gpt-4 & 1347 & 4.324 & 4.143 & 1.869 \\
gpt-4-turbo & 1658 & 4.504 & 4.257 & 1.800 \\
gpt-4o & 1338 & 4.340 & 4.243 & \textbf{1.938} \\
jsl-med-llama-8b & 1337 & 4.196 & 4.016 & 1.818 \\
llama-3-8b & 1374 & 4.097 & 3.832 & 1.845 \\
medlm & 998 & 4.200 & 4.060 & 1.441 \\
medpalm2 & 959 & 3.938 & 3.614 & 1.501 \\
mistral-7b & 1343 & 4.331 & 4.207 & 1.842 \\
phi3-mini-128 & 1361 & 4.212 & 4.139 & 1.882 \\
phi3-mini-4 & 1343 & 4.416 & 4.343 & 1.826 \\
\bottomrule
\end{tabular}
\label{tab:human-eval-consumer-cq}
\end{table*}

\begin{table*}
\caption{Clinician blind evaluations of human and LLM answers showing number of times \textit{Possibility of harm} was rated greater than 1 for all rated MCQs and the percentage change on MCQs human evaluators rated greater than 1 for \textit{Requires African local expertise}. Due to the random order of evaluations and order of model result submissions, models received an unequal number of ratings at the project deadline. How to read this table: for example, of the 1,248 human ratings for gpt-4, 28 responses (2.2\%) were rated as having the possibility of harm. Of the 1,248 ratings, 211 were also rated as requiring African local expertise. Of these 211, 7 MCQ answers (1.9\%) had the possibility of harm, a 15.5\% improvement }
\label{tab:human-eval-harm}
\centering
\begin{tabular}{lrrrrrrr}
\toprule
model & cnt & all & all pct & cnt Afr & all Afr & Afr pct & $\Delta$\% \\
\midrule
medpalm2 & 1 & 56 & 1.786 & 1 & 3 & 33.333 & 1766.349 \\
human & 39 & 1430 & 2.727 & 17 & 314 & 5.414 & 98.533 \\
llama-3-8b & 52 & 1220 & 4.262 & 7 & 174 & 4.023 & -5.608 \\
mistral-7b & 13 & 773 & 1.682 & 5 & 156 & 3.205 & 90.547 \\
claude-3-sonnet & 16 & 857 & 1.867 & 5 & 158 & 3.165 & 69.523 \\
gpt-4-turbo & 18 & 1232 & 1.461 & 5 & 189 & 2.646 & 81.109 \\
phi3-mini-4 & 19 & 1159 & 1.639 & 4 & 177 & 2.260 & 37.889 \\
jsl-med-llama-8b & 35 & 1197 & 2.924 & 4 & 183 & 2.186 & -25.239 \\
gpt-4 & 28 & 1248 & 2.244 & 4 & 211 & 1.896 & -15.508 \\
phi3-mini-128 & 18 & 1127 & 1.597 & 2 & 171 & 1.170 & -26.738 \\
gpt-4o & 23 & 1884 & 1.221 & 3 & 370 & 0.811 & -33.579 \\
gpt-3.5-turbo & 16 & 1202 & 1.331 & 1 & 166 & 0.602 & -54.771 \\
Meta-Llama-3-8B-Instruct & 1 & 166 & 0.602 & 0 & 11 & 0.000 & - \\
Llama3-OpenBioLLM-8B & 2 & 133 & 1.504 & 0 & 5 & 0.000 & - \\
Mixtral-8x7B-Instruct-v0.1 & 1 & 235 & 0.426 & 0 & 9 & 0.000 & - \\
claude-3-haiku & 0 & 83 & 0.000 & 0 & 3 & 0.000 & - \\
claude-3-opus & 0 & 79 & 0.000 & 0 & 4 & 0.000 & - \\
gemini-ultra & 0 & 36 & 0.000 & 0 & 5 & 0.000 & - \\
Llama3-OpenBioLLM-70B & 3 & 169 & 1.775 & 0 & 7 & 0.000 & - \\
gemini-pro & 0 & 40 & 0.000 & 0 & 0 & 0.000 & - \\
Meta-Llama-3-70B-Instruct & 0 & 150 & 0.000 & 0 & 8 & 0.000 & - \\
medlm & 1 & 29 & 3.448 & 0 & 2 & 0.000 & - \\
\bottomrule
\end{tabular}
\end{table*}

\begin{table*}
\caption{Clinician blind evaluations of human and LLM answers showing hallucinations, i.e. number of times \textit{Includes irrelevant, wrong, or extraneous information} was rated greater than 1 for all rated MCQs and the percentage change on MCQs human evaluators rated greater than 1 for \textit{Requires African local expertise}.}
\label{tab:human-eval-hallucinations}
\centering
\begin{tabular}{lrrrrrrr}
\toprule
model & cnt & all & all \% & cnt Afr & all Afr & Afr \% & $\Delta$\% \\
\midrule
llama-3-8b & 117 & 1220 & 9.590 & 11 & 174 & 6.322 & -34.077 \\
jsl-med-llama-8b & 97 & 1197 & 8.104 & 9 & 183 & 4.918 & -39.314 \\
mistral-7b & 47 & 773 & 6.080 & 9 & 156 & 5.769 & -5.115 \\
Meta-Llama-3-8B-Instruct & 10 & 166 & 6.024 & 1 & 11 & 9.091 & 50.913 \\
Llama3-OpenBioLLM-8B & 7 & 133 & 5.263 & 1 & 5 & 20.000 & 280.011 \\
human & 72 & 1430 & 5.035 & 16 & 314 & 5.096 & 1.212 \\
gpt-4o & 89 & 1884 & 4.724 & 18 & 370 & 4.865 & 2.985 \\
Meta-Llama-3-70B-Instruct & 7 & 150 & 4.667 & 1 & 8 & 12.500 & 167.838 \\
claude-3-sonnet & 36 & 857 & 4.201 & 9 & 158 & 5.696 & 35.587 \\
phi3-mini-4 & 43 & 1159 & 3.710 & 5 & 177 & 2.825 & -23.854 \\
gpt-4 & 45 & 1248 & 3.606 & 7 & 211 & 3.318 & -7.987 \\
gpt-4-turbo & 44 & 1232 & 3.571 & 5 & 189 & 2.646 & -25.903 \\
Llama3-OpenBioLLM-70B & 6 & 169 & 3.550 & 0 & 7 & 0.000 & - \\
gpt-3.5-turbo & 36 & 1202 & 2.995 & 5 & 166 & 3.012 & 0.568 \\
gemini-pro & 1 & 40 & 2.500 & 0 & 0 & 0.000 & - \\
phi3-mini-128 & 26 & 1127 & 2.307 & 5 & 171 & 2.924 & 26.745 \\
claude-3-opus & 1 & 79 & 1.266 & 0 & 4 & 0.000 & - \\
Mixtral-8x7B-Instruct-v0.1 & 2 & 235 & 0.851 & 0 & 9 & 0.000 & - \\
claude-3-haiku & 0 & 83 & 0.000 & 0 & 3 & 0.000 & - \\
gemini-ultra & 0 & 36 & 0.000 & 0 & 5 & 0.000 & - \\
medpalm2 & 0 & 56 & 0.000 & 0 & 3 & 0.000 & - \\
medlm & 0 & 29 & 0.000 & 0 & 2 & 0.000 & - \\
\bottomrule
\end{tabular}
\end{table*}

\begin{table*}
\caption{Clinician blind evaluations of human and LLM answers showing ommissions, i.e. number of times \textit{Omission of relevant info} was rated greater than 1 for all rated MCQs and the percentage change on MCQs human evaluators rated greater than 1 for \textit{Requires African local expertise}.}
\label{tab:human-eval-omission}
\centering
\begin{tabular}{lrrrrrrr}
\toprule
model & cnt & all & all \% & cnt Afr & all Afr & Afr \% & $\Delta$\% \\
\midrule
Llama3-OpenBioLLM-8B & 36 & 133 & 27.068 & 1 & 5 & 20.000 & -26.112 \\
llama-3-8b & 264 & 1220 & 21.639 & 39 & 174 & 22.414 & 3.581 \\
jsl-med-llama-8b & 201 & 1197 & 16.792 & 29 & 183 & 15.847 & -5.628 \\
Llama3-OpenBioLLM-70B & 24 & 169 & 14.201 & 2 & 7 & 28.571 & 101.190 \\
human & 201 & 1430 & 14.056 & 56 & 314 & 17.834 & 26.878 \\
gpt-4 & 146 & 1248 & 11.699 & 18 & 211 & 8.531 & -27.079 \\
gpt-4-turbo & 120 & 1232 & 9.740 & 19 & 189 & 10.053 & 3.214 \\
phi3-mini-128 & 107 & 1127 & 9.494 & 16 & 171 & 9.357 & -1.443 \\
mistral-7b & 73 & 773 & 9.444 & 15 & 156 & 9.615 & 1.811 \\
gpt-3.5-turbo & 113 & 1202 & 9.401 & 17 & 166 & 10.241 & 8.935 \\
phi3-mini-4 & 108 & 1159 & 9.318 & 20 & 177 & 11.299 & 21.260 \\
claude-3-sonnet & 72 & 857 & 8.401 & 21 & 158 & 13.291 & 58.207 \\
gpt-4o & 139 & 1884 & 7.378 & 25 & 370 & 6.757 & -8.417 \\
medpalm2 & 4 & 56 & 7.143 & 1 & 3 & 33.333 & 366.653 \\
gemini-pro & 2 & 40 & 5.000 & 0 & 0 & 0.000 & - \\
Meta-Llama-3-70B-Instruct & 4 & 150 & 2.667 & 0 & 8 & 0.000 & - \\
Meta-Llama-3-8B-Instruct & 3 & 166 & 1.807 & 0 & 11 & 0.000 & - \\
claude-3-opus & 1 & 79 & 1.266 & 0 & 4 & 0.000 & - \\
Mixtral-8x7B-Instruct-v0.1 & 2 & 235 & 0.851 & 1 & 9 & 11.111 & 1205.640 \\
claude-3-haiku & 0 & 83 & 0.000 & 0 & 3 & 0.000 & - \\
gemini-ultra & 0 & 36 & 0.000 & 0 & 5 & 0.000 & - \\
medlm & 0 & 29 & 0.000 & 0 & 2 & 0.000 & - \\
\bottomrule
\end{tabular}
\end{table*}

\begin{table*}
\caption{Specialty Human Eval: Clinician ratings for all model responses across specialties showing counts of each specialty, mean of correctness rating, and MCQ counts with percentages where model responses demonstrated the possibility of harm, hallucinations, and omission of relevant information, i.e. ratings greater than 1 for relevant criteria. Bold represents bottom 2 specialties with worst performance across all models for each evaluation axis}
\centering
\begin{tabular}{lrrlll}
\toprule
specialty & count & correct & \# harm (\%) & \# hallucination (\%) & \# omission (\%) \\
\midrule
Obstetrics and Gynecology & 482 & 4.620 & 10 (2.07) & 29 (6.02) & 43 (8.92) \\
Internal Medicine & 323 & 4.635 & 9 (2.79) & 19 (5.88) & 24 (7.43) \\
Infectious Disease & 319 & 4.680 & 3 (0.94) & 5 (1.57) & 9 (2.82) \\
Cardiology & 280 & 4.736 & 4 (1.43) & 11 (3.93) & 8 (2.86) \\
Pediatrics & 274 & 4.661 & 2 (0.73) & 10 (3.65) & 18 (6.57) \\
Endocrinology & 254 & 4.701 & 3 (1.18) & 13 (5.12) & 8 (3.15) \\
Neurology & 217 & 4.765 & 3 (1.38) & 4 (1.84) & 8 (3.69) \\
Other & 215 & 4.521 & 3 (1.40) & 6 (2.79) & 11 (5.12) \\
Gastroenterology & 177 & 4.616 & 6 (3.39) & 13 (7.34) & 17 (9.60) \\
Hematology & 154 & 4.766 & 2 (1.30) & 1 (0.65) & 7 (4.55) \\
General Surgery & 144 & 4.597 & \textbf{8 (5.56)} & \textbf{12 (8.33)} & 13 (9.03) \\
Pathology & 108 & 4.731 & 2 (1.85) & 1 (0.93) & 9 (8.33) \\
Family Medicine & 107 & 4.748 & 0 (0.00) & 5 (4.67) & 6 (5.61) \\
Orthopedic Surgery & 97 & 4.680 & 0 (0.00) & 7 (7.22) & 6 (6.19) \\
Pulmonary Medicine & 94 & 4.755 & 3 (3.19) & 4 (4.26) & \textbf{10 (10.64)} \\
Ophthalmology & 88 & 4.602 & 2 (2.27) & 3 (3.41) & \textbf{10 (11.36)} \\
Oncology & 87 & 4.678 & 2 (2.30) & 4 (4.60) & 4 (4.60) \\
Nephrology & 79 & 4.570 & 1 (1.27) & 6 (7.59) & 8 (10.13) \\
Emergency Medicine & 71 & 4.662 & 0 (0.00) & 4 (5.63) & 4 (5.63) \\
Anesthesiology & 69 & 4.725 & \textbf{3 (4.35)} & 2 (2.90) & 5 (7.25) \\
Urology & 56 & 4.714 & 0 (0.00) & 2 (3.57) & 4 (7.14) \\
Psychiatry & 54 & 4.667 & 1 (1.85) & 3 (5.56) & 3 (5.56) \\
Dermatology & 50 & 4.700 & 1 (2.00) & 1 (2.00) & 2 (4.00) \\
Neurosurgery & 48 & 4.438 & 1 (2.08) & 2 (4.17) & 5 (10.42) \\
Allergy and Immunology & 43 & 4.605 & 1 (2.33) & 2 (4.65) & 3 (6.98) \\
Otolaryngology & 35 & 4.743 & 1 (2.86) & 0 (0.00) & 2 (5.71) \\
Radiology & 23 & 4.478 & 0 (0.00) & \textbf{2 (8.70)} & 1 (4.35) \\
Medical Genetics & 18 & 5.000 & 0 (0.00) & 0 (0.00) & 0 (0.00) \\
Plastic Surgery & 17 & 4.706 & 0 (0.00) & 1 (5.88) & 1 (5.88) \\
Rheumatology & 16 & 4.625 & 0 (0.00) & 1 (6.25) & 1 (6.25) \\
Geriatrics & 11 & 5.000 & 0 (0.00) & 0 (0.00) & 0 (0.00) \\
\bottomrule
\end{tabular}
\label{tab:human-eval-specialties}
\end{table*}

\end{document}